%% file: templateArxiv.tex
\documentclass[twocolumn]{article}

\usepackage{PRIMEarxiv}

\usepackage[utf8]{inputenc} % allow utf-8 input
\usepackage[T1]{fontenc}    % use 8-bit T1 fonts
\usepackage{hyperref}       % hyperlinks
\usepackage{url}            % simple URL typesetting
\usepackage{booktabs}       % professional-quality tables
\usepackage{amsfonts}       % blackboard math symbols
\usepackage{nicefrac}       % compact symbols for 1/2, etc.
\usepackage{microtype}      % microtypography
\usepackage{lipsum}
\usepackage{fancyhdr}       % header
\usepackage{graphicx}       % graphics
\usepackage[square, numbers]{natbib}
\usepackage{amssymb}% http://ctan.org/pkg/amssymb
\usepackage{pifont}% http://ctan.org/pkg/pifont
\newcommand{\cmark}{\ding{51}}
\newcommand{\xmark}{\ding{55}}
\usepackage{orcidlink}
\usepackage{array}
\newcolumntype{M}[1]{>{\centering\arraybackslash}m{#1}}
\newcolumntype{P}[1]{>{\centering\arraybackslash}p{#1}}
\newcolumntype{N}{@{}m{0pt}@{}}
\usepackage{float}
\usepackage{multirow}
\usepackage{comment}
\usepackage[acronym]{glossaries}
\glsdisablehyper
\input{acrons}

\usepackage{hyperref} 
\hypersetup{backref=true, 
    pagebackref=true,               
    hyperindex=true,                
    colorlinks=true,                
    breaklinks=true,                
    urlcolor=black,                
    linkcolor=blue,                
    bookmarks=true,                 
    bookmarksopen=false,
    filecolor=black,
    citecolor=blue
}

\title{Markerless Multi-view 3D Human Pose Estimation: a survey}
\author{
    \begin{tabular}{c c c}
        Ana Filipa Rodrigues Nogueira \orcidlink{0000-0002-9413-3300}\textsuperscript{1,2} & Hélder P. Oliveira \orcidlink{0000-0002-6193-8540}\textsuperscript{1,3} & Luís F. Teixeira \orcidlink{0000-0002-4050-7880}\textsuperscript{1,2} \\
        \texttt{\small ana.f.rodrigues@inesctec.pt} & \texttt{\small helder.f.oliveira@inesctec.pt} & \texttt{\small luisft@fe.up.pt} 
    \end{tabular} \\
    \vspace{10pt}
    \\
    {\textsuperscript{1} Instituto de Engenharia de Sistemas e Computadores, Tecnologia e Ciência (INESC TEC), Rua Dr. Roberto Frias,}\\{4200-465, Porto, Portugal}\\
    {\textsuperscript{2} Faculdade de Engenharia da Universidade do Porto (FEUP), Rua Dr. Roberto Frias, 4200-465, Porto, Portugal}\\
    {\textsuperscript{3} Faculdade de Ciências da Universidade do Porto (FCUP), Rua do Campo Alegre, 1021-1055, Porto, Portugal}\\
}

\begin{document}

\twocolumn[
\begin{@twocolumnfalse}

\maketitle

\begin{abstract}
3D human pose estimation involves reconstructing the human skeleton by detecting the body joints. Accurate and efficient solutions are required for several real-world applications including animation, human-robot interaction, surveillance, and sports. However, challenges such as occlusions, 2D pose mismatches, random camera perspectives, and limited 3D labelled data have been hampering the models' performance and limiting their deployment in real-world scenarios. The higher availability of cameras has led researchers to explore multi-view solutions to take advantage of the different perspectives to reconstruct the pose.

Most existing reviews have mainly focused on monocular 3D human pose estimation, so a comprehensive survey on multi-view approaches has been missing since 2012.  
According to the reviewed articles, the majority of the existing methods are fully-supervised approaches based on geometric constraints, which are often limited by 2D pose mismatches. To mitigate this, researchers have proposed incorporating temporal consistency or depth information. Alternatively, working directly with 3D features has been shown to completely overcome this issue, albeit at the cost of increased computational complexity. Additionally, models with lower levels of supervision have been identified to help address challenges such as annotated data scarcity and generalisation to new setups. 
Therefore, no method currently addresses all challenges associated with 3D pose reconstruction, and a trade-off between complexity and performance exists. Further research is needed to develop approaches capable of quickly inferring a highly accurate 3D pose with bearable computation cost. Techniques such as active learning, low-supervision methods, temporal consistency, view selection, depth information estimation, and multi-modal approaches are strategies to keep in mind when developing a new methodology to solve this task.
\end{abstract}

% keywords can be removed

\end{@twocolumnfalse}]

\input{sections/Introduction}
\input{sections/Datasets}
\input{sections/metrics}

\input{sections/Multi-view}
\input{sections/WiFi_RF_signals}
\input{sections/Discussion}

\input{sections/Challenges_and_Opportunities}

\vspace{-0.1cm}\section*{Acknowledgments}\vspace{-0.2cm}
This work is co-financed by Component 5 - Capitalization and Business Innovation of core funding for Technology and Innovation Centres (CTI), integrated in the Resilience Dimension of the Recovery and Resilience Plan within the scope of the Recovery and Resilience Mechanism (MRR) of the European Union (EU), framed in the Next Generation EU, for the period 2021–2026, with reference 21 and by National Funds through the Portuguese funding agency, FCT-Foundation for Science and Technology Portugal, a PhD Grant Number 2023.02851.BD.

%% \appendix~
\input{sections/Benchmarking}

%Bibliography
\bibliographystyle{plainnat}

\input{output.tex}
\end{document}

%% file: acrons.tex
\makeglossaries
\newacronym{dl}{DL}{Deep Learning}
\newacronym{ml}{ML}{Machine Learning}
\newacronym{csi}{CSI}{Channel State Information}
\newacronym{rf}{RF}{Radio-Frequency}
\newacronym{lidar}{LiDAR}{Light Detection And Ranging}
\newacronym{prisma}{PRISMA}{Preferred Reporting Items for Systematic Reviews and Meta-Analyses}
\newacronym{hpe}{HPE}{Human Pose Estimation}
\newacronym{imus}{IMUs}{Inertial Measurement Units}
\newacronym{rgb-d}{RGB-D}{Red Green Blue-Depth}
\newacronym{pcp}{PCP}{Percentage of Correct Parts}
\newacronym{mpjpe}{MPJPE}{Mean Per Joint Position Error}
\newacronym{pck}{PCK}{Percentage of Correct Keypoints}
\newacronym{mrpe}{MRPE}{Mean of the Root Position Error}
\newacronym{mpjae}{MPJAE}{Mean Per Joint Angle Error}
\newacronym{oks}{OKS}{Object Keypoint Similarity}
\newacronym{ps}{PS}{Pictorial Structures}
\newacronym{3dps}{3DPS}{3D Pictorial Structures}
\newacronym{svm}{SVM}{Support Vector Machine}
\newacronym{fmp}{FMP}{Flexible Mixtures of Parts}
\newacronym{apf}{APF}{Annealed Particle Filter}
\newacronym{crf}{CRF}{Conditional Random Field}
\newacronym{gmm}{GMM}{Gaussian Mixture Model}
\newacronym{dnn}{DNN}{Deep Neural Network}
\newacronym{dlt}{DLT}{Direct Linear Transform}
\newacronym{map}{MAP}{Maximum A Posteriori}
\newacronym{cpn}{CPN}{Cuboid Proposal Network}
\newacronym{prn}{PRN}{Pose Regression Network}
\newacronym{cnn}{CNN}{Convolutional Neural Network}
\newacronym{vtp}{VTP}{Volumetric Transformer Pose Estimator}
\newacronym{ssa}{SSA}{Sparse Sinkhorn Attention}
\newacronym{tof}{ToF}{Time-of-Flight}
\newacronym{umpm}{UMPM}{Utrecht Multi-Person Motion}
\newacronym{mvor}{MVOR}{Multi-View Operating Room}
\newacronym{ap}{AP}{Average Precision}
\newacronym{mAP}{mAP}{Mean Average Precision}
\newacronym{tp}{TP}{True Positive}
\newacronym{fp}{FP}{False Positive}
\newacronym{dpd}{DPD}{Depth-wise Projection Decay}
\newacronym{edn}{EDN}{Encoder-Decoder Network}
\newacronym{auc}{AUC}{Area Under the Curve}
\newacronym{pa-mpjpe}{PA-MPJPE}{Procrustes Alignment MPJPE}
\newacronym{pca}{PCA}{Principal Component Analysis}
\newacronym{ifid}{IFID}{Inter-Frame and Intra-Frame Desynchronized Dataset}
\newacronym{pp}{pp}{percentual points}

%% file: sections/Introduction.tex
\section{Introduction}\label{sec:intro}
3D human pose estimation aims to reconstruct the body configuration of all persons in a scene. Finding solutions for this task is essential for numerous applications ranging from human-robot interaction \cite{human_robot_interaction2020Xu}, animation \cite{Hwang2023rgbd_animation}, gaming, action recognition \cite{Liu2021dualview}, rehabilitation assessments, surveillance systems \cite{fast_and_robust2022}, sports \cite{Mustafa2022_4DTemporallyCoherent,bridgeman2019sports}, live broadcasts \cite{Fan2021Multiagent}, human-computer interaction, such as recognising sign language \cite{how2sign}, among many others. As an example of application, the work of Mustafa et al. \cite{Mustafa2022_4DTemporallyCoherent} showed that the reconstruction of the 3D pose in a multi-view setting helped the creation of a method for 4D dynamic scene understanding with numerous interacting individuals, such as sports games.

Therefore, it is necessary to develop solutions that can solve this task effectively and efficiently, without the need to use markers. Because the use of markers imposes many restrictions on the application scenarios in which the methods can be used. Furthermore, placing markers over clothing can lead to incorrect readings of the person's keypoints due to possible displacement of a marker during the performance of movements \cite{humaneva}.

Methods for estimating the 2D pose have been widely explored, however, those only allowed the reconstruction of a surface pose. Thus, to capture the volume of the body of the person, it is necessary to determine the 3D pose. Nonetheless, since there have been more advances in 2D pose estimation methods, many 3D estimation algorithms, use the 2D pose estimations for each view to reconstruct the 3D pose.

The use of multiple views helps to capture the whole body geometry making it easier and more suitable for 3D pose estimation than monocular methodologies \cite{tu2020voxelpose}. However, the occurrence of occlusions, poor camera calibration, lack of 3D annotated data, similarities or variations in human appearance, and difficulties in associating the multiple views and generalise to new perspectives are some challenges methods developed for multi-view systems have to overcome to accurately estimate the 3D pose \cite{bridgeman2019sports, Integrating_Multiple_Uncalibrated_Views}.

\subsection{Previous literature reviews}
Solutions to accurately and efficiently estimate the 3D human pose have been widely explored over the years. Thus, there are numerous surveys summarizing and evaluating most of the existing works. 

Initially, most of the methodologies were developed for single-view inputs. Therefore, there are several surveys addressing 3D pose estimation based on monocular images \cite{monoculardlsurvey_2020,SONG2021_survey,JoaoM_2step,monocular_survey2020,recentmonocular_survey2022,overview3dpose}. The main findings are that, even though deep learning has brought considerable improvements to the estimation of the 3D pose, occlusions, crowded scenarios, and a lack of datasets that can realistically mimic real-world settings still limit the models' performance and restrain their real-world employment. For future research, it is suggested the use of transfer learning \cite{JoaoM_2step}, synthetic data \cite{JoaoM_2step,recentmonocular_survey2022}, models with little supervision \cite{recentmonocular_survey2022}, or complementary information by, for example, exploiting a multi-modal approach \cite{SONG2021_survey,recentmonocular_survey2022}, incorporate global and local context to obtain more distinguishing characteristics \cite{monoculardlsurvey_2020}, explore the interactions between the individuals and the scene or the objects \cite{monocular_survey2020,recentmonocular_survey2022,overview3dpose} or add the use of temporal information \cite{JoaoM_2step}. In addition, multi-view geometry is pointed out as a solution for data scarcity and the ambiguities in monocular 3D pose estimation \cite{monocular_survey2020,recentmonocular_survey2022,overview3dpose}.

The increased availability of multi-camera setups has prompted further research on the benefits of having various camera perspectives. As a result, several surveys, besides analysing single-view methods, also cover multi-view approaches \cite{electronics_survey_supervision,LIU2015_parsing,DESMARAIS2021103275,SARAFIANOS20161,WANG2021103225,hand_survey,BENGAMRA2021104282,deeplearning_2023,vision_survey,Kumar2022}.  
The literature review of Holte et al. \cite{Holte_survey} is the only one found, solely dedicated to multi-view methodologies. 
According to the analysed surveys, most of the problems observed in monocular settings are the same for multi-view, namely, the lack of annotated in-the-wild datasets or with rare poses \cite{LIU2015_parsing,SARAFIANOS20161,BENGAMRA2021104282,DESMARAIS2021103275,vision_survey,WANG2021103225,deeplearning_2023,hand_survey}, the difficulty in identifying the several poses in crowded scenarios, and the occlusions \cite{electronics_survey_supervision,Kumar2022,deeplearning_2023,vision_survey,Holte_survey}. Even though, overcoming the occlusions problem benefits from the different view perspectives, it still remains a challenge in some situations. Besides, the increase in the number of viewpoints leads to the need for more complex models to be able to deal with the information acquired from all perspectives. This translates into slower inference runtime, forcing a balance between complexity and speed \cite{BENGAMRA2021104282,Kumar2022,deeplearning_2023,vision_survey}. Furthermore, the bad image quality in terms of focus or blur \cite{deeplearning_2023,Kumar2022,vision_survey,Holte_survey}, the inaccuracies of 2D pose estimations which affect the models that use 2D-3D lifting \cite{vision_survey,Holte_survey} are also, factors negatively influencing the quality of the predictions. So, there are still a lot of opportunities for improvement, such as considering the human interactions either with other humans or the surrounding environment \cite{SARAFIANOS20161,WANG2021103225,deeplearning_2023,Holte_survey}, the increase of the generality in order for the same method to be suitable across multiple applications and independent of the presented viewpoints \cite{Kumar2022,vision_survey,Holte_survey}. Also, several surveys recommend the use of neural architecture search for a better and more efficient design of neural networks \cite{WANG2021103225,Kumar2022,deeplearning_2023,vision_survey}, taking advantage of temporal information, even the adaptation of methods from monocular to multi-view \cite{Kumar2022,DESMARAIS2021103275}, the use of transfer learning \cite{LIU2015_parsing} and data augmentation \cite{deeplearning_2023}. Finally, to overcome the scarcity of annotated datasets, several strategies are suggested like using unsupervised, semi-supervised or weakly-supervised models \cite{LIU2015_parsing,WANG2021103225}, or the use of active learning to alleviate the human workload \cite{hand_survey}.

Moreover, some surveys briefly mention multi-modal approaches, mainly, combining vision and \gls{imus} sensors but also, depth sensors, point clouds or \gls{rf}, which allow to obtain more accurate estimations, demonstrating the benefits of having complementary information \cite{DESMARAIS2021103275,deeplearning_2023,vision_survey}.

Therefore, the scope of the present survey is in markerless methods, so, it is only going to be considered methods that do not require people to have attached sensors. Also, the literature lacks a review focus on methods which used various cameras to determine the pose. Most of the surveys that address these methods only present a relatively brief section, as can be seen in Table \ref{tab:surveys_summary}.
Thus, this literature review will solely focus on 3D Human Pose Estimation based on data from multiple cameras and also, acknowledge works that use different types of sensors, such as \gls{rgb-d} cameras, \gls{tof} cameras or wireless devices, to obtain the 3D pose in a multi-view environment.

\begin{table*}[!htb]
\centering
\scriptsize
\caption{Table of a comparative analysis between the surveys on 3D pose estimation}
\begin{tabular}{m{5.5cm}M{0.5cm}M{1.2cm}M{1.2cm}M{1cm}M{1cm}M{1.8cm}M{1cm}}
\toprule
\multicolumn{1}{c}{\textbf{Papers}} & \textbf{Year} & \textbf{Monocular view methods} & \textbf{Multi-view methods} & \textbf{Wi-Fi, RF or other sensors} & \textbf{Datasets} & \textbf{Benchmarking Performance} & \textbf{Metrics} \\ \midrule
A survey on monocular 3D human pose estimation \cite{monocular_survey2020} & 2020 & \cmark & \xmark & \xmark & \cmark & \cmark & \cmark \\ \midrule
Monocular human pose estimation: A survey of deep learning-based methods \cite{monoculardlsurvey_2020} & 2020 & \cmark & \xmark & \xmark & \cmark & \cmark & \cmark \\ \midrule
Human pose estimation and its application to action recognition: A survey \cite{SONG2021_survey} & 2021 & \cmark & \xmark & \xmark & \cmark & \cmark & \xmark \\ \midrule
A Survey of Recent Advances on Two-Step 3D Human Pose Estimation \cite{JoaoM_2step} & 2022 & \cmark & \xmark & \xmark & \cmark & \cmark & \xmark \\ \midrule
Recent Advances of Monocular 2D and 3D Human Pose Estimation: A Deep Learning Perspective \cite{recentmonocular_survey2022} & 2022 & \cmark & \xmark & \xmark & \cmark & \cmark & \cmark \\ \midrule
Efficient Annotation and Learning for 3D Hand Pose Estimation: A Survey \cite{hand_survey} & 2023 & \cmark & \cmark \text{\scriptsize\hspace{0.05cm}(19 works)} 
& \xmark & \xmark & \xmark & \xmark \\ \midrule
{Human pose estimation using deep learning: review, methodologies, progress and future research directions \cite{Kumar2022}} & 2022 & \cmark & \cmark \text{\scriptsize\hspace{0.05cm}(9 works)} 
& \xmark & \cmark & \xmark & \cmark \\ \midrule
A survey of human pose estimation: The body parts parsing based methods \cite{LIU2015_parsing} & 2015 & \cmark & \cmark \text{\scriptsize\hspace{0.05cm}(6 works)} 
& \xmark & \cmark & \cmark & \xmark \\ \midrule
3D Human pose estimation: A review of the literature and analysis of covariates \cite{SARAFIANOS20161} & 2016 & \cmark & \cmark \text{\scriptsize\hspace{0.05cm}(11 works)} 
& \xmark & \cmark & \cmark & \cmark \\ \midrule
Deep 3D human pose estimation: A review \cite{WANG2021103225} & 2021 & \cmark & \cmark \text{\scriptsize\hspace{0.05cm}(16 works)} 
& \xmark & \cmark & \cmark & \cmark \\ \midrule
Deep Learning Methods for 3D Human Pose Estimation under Different Supervision Paradigms: A Survey \cite{electronics_survey_supervision} & 2021 & \cmark & \cmark \text{\scriptsize\hspace{0.05cm}(11 works)} 
& \xmark & \cmark & \cmark & \cmark \\ \midrule
A review of 3D human pose estimation algorithms for markerless motion capture \cite{DESMARAIS2021103275} & 2021 & \cmark & \cmark \text{\scriptsize\hspace{0.05cm}(7 works)} 
& \xmark & \cmark & \cmark & \cmark \\ \midrule
{A review of deep learning techniques for 2D and 3D human pose estimation \cite{BENGAMRA2021104282}} & 2021 & \cmark & \cmark \text{\scriptsize\hspace{0.05cm}(9 works)} 
& \xmark & \cmark & \cmark & \cmark \\ \midrule
Overview of 3D Human Pose Estimation \cite{overview3dpose} & 2023 & \cmark & \cmark \text{\scriptsize\hspace{0.05cm}(17 works)} 
& \xmark & \cmark & \cmark & \cmark \\ \midrule
Deep Learning-Based Human Pose Estimation: A Survey \cite{deeplearning_2023} & 2023 & \cmark & \cmark \text{\scriptsize\hspace{0.05cm}(26 works)} & \cmark & \cmark & \cmark & \cmark \\ \midrule
{Vision-Based Human Pose Estimation via Deep Learning: A Survey \cite{vision_survey}} & 2023 & \cmark & \cmark \text{\scriptsize\hspace{0.05cm}(8 works)} & \cmark & \cmark & \cmark & \cmark \\ \midrule
Human pose estimation and activity recognition from multi-view videos: Comparative explorations of recent developments \cite{Holte_survey} & 2012 & \xmark & \cmark \text{\scriptsize\hspace{0.05cm}(24 works)} & \xmark & \cmark & \cmark & \xmark \\ \midrule
\textbf{Markerless Multi-view 3D Human Pose Estimation: a survey (Ours)} & \textbf{2024} & \textbf{\xmark} & \textbf{\cmark \text{\scriptsize\hspace{0.05cm}(57 works)}} & \textbf{\cmark} & \textbf{\cmark} & \textbf{\cmark} & \textbf{\cmark} \\ \bottomrule
\end{tabular}\label{tab:surveys_summary}
\end{table*}

\subsection{Search Process} 
\begin{figure}[!htb]
 \centering
 \includegraphics[width=0.9\linewidth]{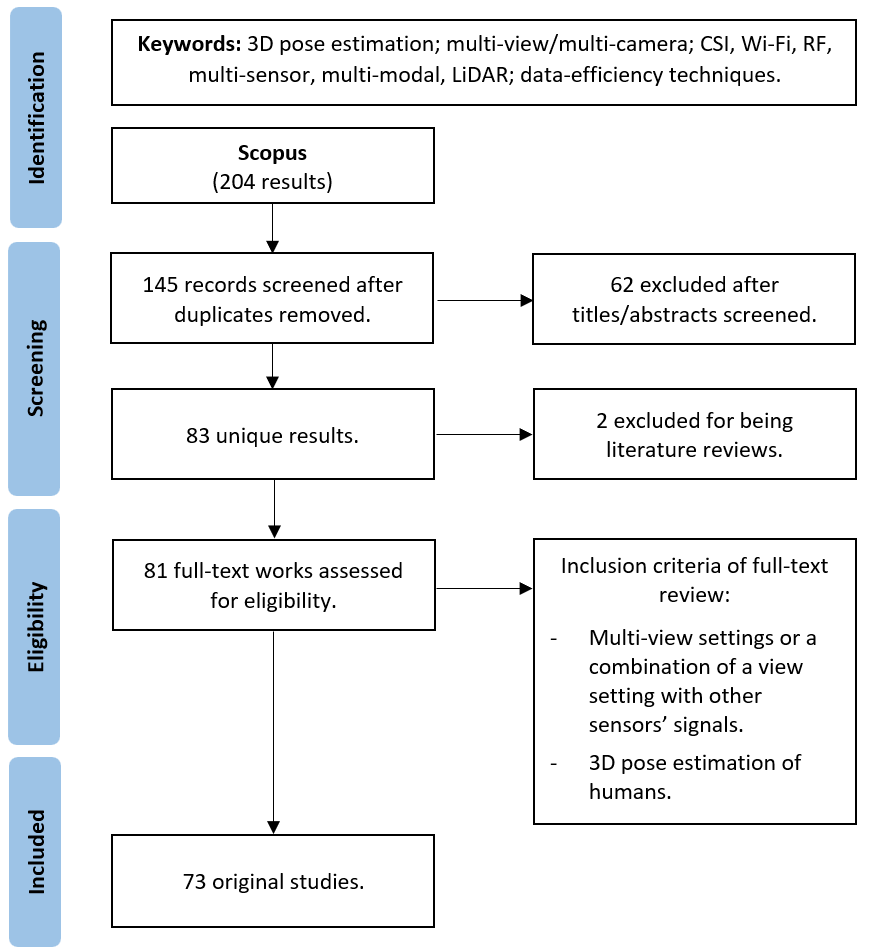}
 \caption{\gls{prisma} diagram of the conducted research process (adapted from \cite{PRISMA_ref}).}
 \label{fig:prisma}
\end{figure}

The research process, Figure \ref{fig:prisma}, was conducted in September 2023 and updated in June 2024, using the search engine for scientific articles: Scopus. Several combinations of the following keywords were used: 3D pose estimation; multi-view/multi-camera; \gls{csi}, \gls{rf}, Wi-Fi, multi-sensor, multi-modal, \gls{lidar}; data-efficiency techniques. 

\begin{figure}[!htb]
 \centering
 \includegraphics[width=\linewidth]{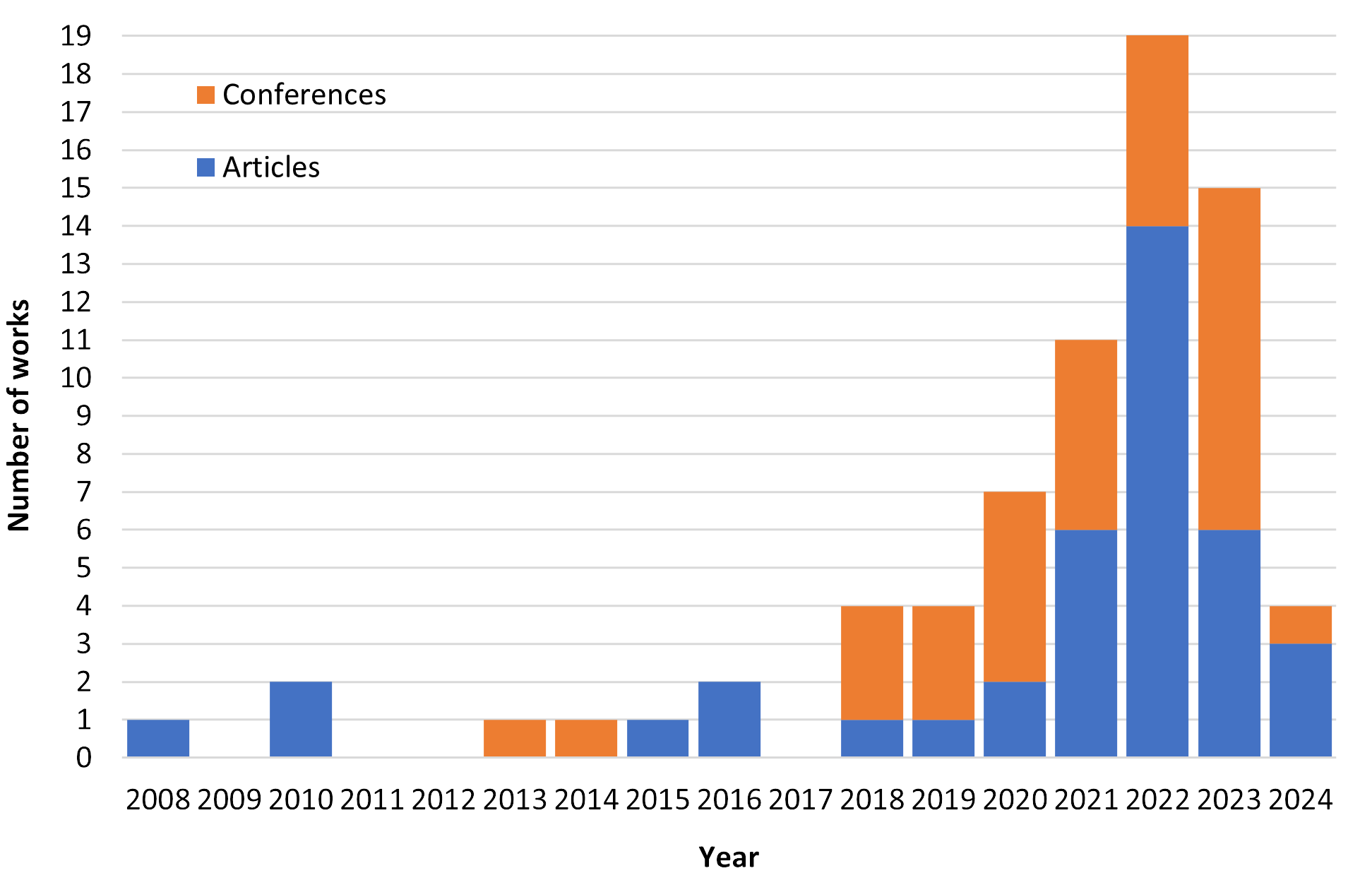}
 \caption{Evolution of the published works found in Scopus database for markerless 3D pose estimation in multi-camera settings or by using other types of sensors such as \gls{rgb-d} cameras, \gls{tof} cameras or wireless sensors.}
 \label{fig:evolution_graph}
\end{figure}

This search process resulted in 204 works, and the following criteria were then applied to filter the results:
\begin{enumerate}
 \item Remove duplicates
 \item Exclude the articles that are entirely unrelated to the subject by analysing the title and abstract.
 \item Include only peer-reviewed works written in English.
 \item Analyse the body of the text and include only the relevant works to the study at hand. 
\end{enumerate}
After applying the above criteria, it resulted in 73 works that were fully revised and used to construct this literature review. In Figure \ref{fig:evolution_graph} is possible to see the growing interest in this area.

The rest of this literature review is organised in the following way: Chapter \ref{sec:datasets} - Datasets, Chapter \ref{sec:metrics} - Evaluation Metrics, Chapter \ref{sec:multiview} - Multi-view approaches, Chapter \ref{sec:multimodal} - Multi-modal approaches, Chapter \ref{sec:discussion} - Discussion, Chapter \ref{sec:conclusion} - Conclusion. 

%% file: sections/Datasets.tex
\section{Datasets}\label{sec:datasets}
The most used datasets to train and assess the performance of 3D Human Pose Estimation in multi-view settings are Human3.6M \cite{h36m_pami}, Campus \cite{Belagiannis2014}, Shelf \cite{Belagiannis2014} and CMU Panoptic \cite{CMU_Panoptic}. 

\textbf{Human3.6M} \cite{h36m_pami} is a commonly used dataset for estimating the 3D pose of a single person. The data was collected in a laboratory with 4 digital video cameras positioned in the corners, 1 \gls{tof} on top of one digital camera, and 10 motion cameras spread throughout the walls, with 4 on each side and 2 in the middle of the horizontal edge. 
The data included 15 distinct actions: discussion, greeting, posing, walking, eating, sitting and taking photographs, among others. These actions were enacted by a total of 11 unique individuals, encompassing both male and female participants. Out of these 11 individuals, two males and two females were exclusively included in the testing set.

\textbf{Campus} \cite{Belagiannis2014} consists of three persons interacting and being recorded by three cameras in an outdoor environment. Annotations for the body joints of the three individuals were created for cameras 1 and 2. Subsequently, these annotations were triangulated and projected to generate the annotations for the third perspective.

\textbf{Shelf} \cite{Belagiannis2014} comprises four people disassembling a shelf in an indoor setting which is being recorded by five cameras. Due to the consistent occlusion of one participant (Actor 4) in the majority of the frames, the pose estimation corresponding to that person is usually disregarded during the evaluation. The 3D ground-truth annotations were generated by triangulating the body joint annotations from the 2$^{nd}$, 3$^{rd}$, and 4$^{th}$ camera perspectives.

\textbf{CMU Panoptic} \cite{CMU_Panoptic} is the dataset with the greatest number of perspectives available, providing a total of 521 views (480 VGA, 31 HD and 10 Kinetic cameras). The data was acquired in a controlled environment and includes both actions performed in groups and individually. Participants were organised into groups of a maximum of 8 elements to participate in social activities such as games like Ultimatum or Haggling, group discussions, musical performances, or dancing.

Other datasets were captured with a specific application scenario in mind, such as KTH Multiview Football \cite{kth_football_i, kth_football_ii}, which allows tracking people on a football field, or MPII Cooking \cite{mp_cooking_ii}, to track people during their cooking process. Table \ref{tab:datasets} summarises the existing datasets and their main characteristics. Table \ref{tab:h36m_results}, Table \ref{tab:campus_shelf_results} and Appendix \ref{sec:benchmarking} present the benchmarking for the several datasets used in the revised works.

\begin{table*}[!htb]
\centering
\scriptsize
\caption{Summary of the found datasets for Multi-view 3D Human Pose Estimation.}
\begin{tabular}{m{2.4cm}M{0.6cm}M{1.5cm}M{1.2cm}M{1.2cm}m{7cm}}
\toprule
\textbf{Datasets} & \textbf{Year} & \begin{tabular}[c]{@{}c@{}}\textbf{Size}\\ \textbf{(Nº frames)}\end{tabular} & \textbf{Nº cameras} & \textbf{Nº subjects} & \textbf{Characteristics}\\
\midrule
HumanEva-I \cite{humaneva} & 2010 & 40 060 & 7 & 4 & Single-person dataset that includes 6 distinct actions: random hand movements, boxing, playing with a ball by tossing and catching, walking, running and a combined sequence in which the subject begins walking, then runs and at the end, balances on each foot alternately. The ground-truth data was obtained using a MoCap system with 6 cameras. \\\midrule
HumanEva-II \cite{humaneva} & 2010 & 2 460 & 4 & 2 & Single-person performing an extended sequence of actions compared to HumanEva-I. The ground-truth motion was captured using a MoCap system with 12 cameras. This is only a test set, the models are supposed to be trained and validated with the data from HumanEva-I. \\\midrule
\gls{umpm} \cite{umpm_dataset} & 2011 & 400 000 & 4 & 30 & Consists of 9 different scenarios recorded with 1 to 4 people in the scene. Ground-truth was captured with 14 Vicon MoCap cameras. \\\midrule
KTH Multiview Football I \cite{kth_football_i} & 2012 & 257 & 3 & 2 & Professional football players during a match of the Allsvenskan league. \\\midrule
KTH Multiview Football II \cite{kth_football_ii} & 2013 & 800 & 3 & 2 & Extended version of KTH Multiview Football I. \\\midrule
Human3.6M \cite{h36m_pami} & 2014 & 3 600 000 & 4 & 11 & Only single-person scenarios realizing one of the 17 pre-defined actions. People have IMU sensors attached for a better annotation of the keypoints. \\\midrule
Campus \cite{Belagiannis2014} & 2014 & 2 000 & 3 & 3 & Multiple people interacting in an outdoor environment. \\\midrule
Shelf \cite{Belagiannis2014} & 2014 & 3 200 & 5 & 4 & Multiple people in a room disassembling a shelf. \\\midrule
CMU Panoptic \cite{CMU_Panoptic} & 2015 & 1 500 000 & 521 & 1-8 per frame & Contains scenes with a single person doing a set of movements and scenarios with multiple people engaging in social activities. (521 cameras: 480 VGA camera; 31 HD cameras and 10 Kinect II Sensors). \\\midrule
MPII Cooking 2 \cite{mp_cooking_ii} & 2015 & 2 881 616 & 8 & 30 & Single-person preparing several dishes. \\\midrule
NTU RGB+D \cite{ntu_rgbd} & 2016 & 4 000 000 & 3 & 40 & There are 60 different actions; some were performed with just one person in the scene and some required a multi-person setting. It was collected depth maps, RGB images and videos, 3D joint information, and infrared sequences. \\\midrule
MPI-INF-3DHP \cite{mpi_inf_3dhp} & 2017 & \textgreater{}1 300 000 & 14 & 8 & Single-person performing one of the 8 pre-defined activities. \\\midrule
Total Capture \cite{Totalcapture} & 2017 & 1 892 176 & 8 & 5 & Single-person either walking, acting, doing freestyle movements or a range of motion sequences. All frames have a multi-view video, IMU and Vicon labelling. \\\midrule
PKU-MMD \cite{pku_mmd_dataset} & 2017 & 5 312 580 & 3 & 66 & The subjects performed various actions by themselves and in group. Contains depth maps, RGB images, skeleton joints, infrared sequences, and RGB videos. \\\midrule
WILDTRACK \cite{wildtrack} & 2018 & 36 000 & 7 & 313 & Data captured in the street in front of the ETH Zurich University's main building.\\ \midrule
\gls{mvor} \cite{mvor_dataset} & 2018 & 732 & 3 & 10 & Consists of images recorded during 4 days in an interventional room at the University Hospital of Strasbourg. \\\midrule
NTU RGB+D 120 \cite{ntu_rgbd_120} & 2019 & 8 000 000 & 3 & 106 &  It is an extension of NTU RGB+D and it has 60 more different actions, making a total of 120 actions.\\\midrule
%\end{tabular}
%\label{tab:datasets}
%\end{table*}
%\begin{table*}[!htb]
%\centering
%\small
%\begin{tabular}{m{2.4cm}M{0.6cm}M{1.5cm}M{1.2cm}M{1.2cm}m{7cm}}
%\midrule
How2Sign \cite{how2sign} & 2021 & $\sim$7 503 218 & 3 & 11 & Consists of people doing sign language. Besides images, it also, contains speech, English transcripts, gloss, pose information and depth information. \\\midrule
HuMoMM \cite{HuMoMM_dataset} & 2023 & 262 000 & 5 & 18  &  Includes 30 different actions: 20 performed individually and 10 in group. Contains RGB images and depth images and also, provides multi-modal annotations that comprehend 2D and 3D keypoints, SMPL parameters and action categories. \\\midrule
PKUInfantV \cite{YIN2024_infantpose} & 2024 & $\sim$2 000 000 & 3 & 170 & It comprises 510 videos recorded in a hospital setting, and the movements of the infant were categorized as either normal or abnormal writhing motions by a physician. This dataset led to the creation of two other datasets: one consisting of 15 924 annotated images obtained from 420 videos in the PKUInfantV dataset, and the other being a downsampled version featuring solely the segments of the videos in which the infant exhibited movement.    \\ 
\bottomrule
\end{tabular}
\label{tab:datasets}
\end{table*}

%% file: sections/metrics.tex
\section{Evaluation Metrics}\label{sec:metrics}
This section presents the evaluation metrics specific to human pose estimation. The most widely used metrics to evaluate the quality of the 3D pose estimations are \gls{pcp}, \gls{mpjpe}, \gls{pck}, or \gls{ap}. Nonetheless, other less frequently used metrics like \gls{mAP}, \gls{auc}, \gls{mpjae} and \gls{mrpe} are also introduced.\\

\textbf{\gls{pcp}} determines the limb detection accuracy. A limb is considered correct if the distance between the predicted limb ends and the ground-truth limb joint locations is less or equal to a certain value $\alpha$ of the limb's length, (see Eq. \ref{eq:pcp3d}).
\begin{equation}\label{eq:pcp3d}
    \frac{\left|\left|s_p - s_p^{'}\right|\right| + \left|\left|e_p - e_p^{'}\right|\right|}{2} \leq \alpha\left|\left|s_p-e_p\right|\right|
\end{equation}
$s_p$ and $e_p$ are the 3D ground-truth coordinates of the beginning and ending points of the body part $p$.\\
$s_p^{'}$ and $e_p^{'}$ are the corresponding estimated coordinates of the beginning and ending points of the body part $p$.\\
$\alpha$ is the threshold, normally defined as $0.5$.\\

\textbf{\gls{mpjpe}}, also known as 3D error, corresponds to the average Euclidean distance between the estimated joints and the respective true joint location, (see Eq. \ref{eq:mpjpe}).
\begin{equation}\label{eq:mpjpe}
    \text{MPJPE }(S) = \frac{1}{N_S}\sum_{i=1}^{N_S}\left|\left|G_i-P_i\right|\right|_2
\end{equation}
$G_i$ is the ground-truth position of the i-the joint.\\
$P_i$ is the predicted position of the i-the joint.\\
$N_S$ is the number of joints of the $S$ skeleton.\\

Some researchers also report results using a variant of \gls{mpjpe}, \gls{pa-mpjpe}, which uses Procrustes alignment \cite{Gower1975GeneralizedPA} on the ground-truth and estimate joints before calculating MPJPE.\\

\textbf{\gls{pck}}
consists in the \% of points in which the distance between the estimated and the real value is inferior to a defined threshold, (see Eq. \ref{eq:pck}). 
\begin{equation}\label{eq:pck}
    PCK_{k_i} = \frac{1}{N}\sum_{i=1}^N \delta\left(\left|\left|P_k^i-G_k^i\right|\right|_2^2 \leq t_j\right)
\end{equation}
$PCK_{k_i}$ is the \gls{pck} value of the i-the keypoint of the k-the skeleton.\\
$t_j$ is the j-the defined threshold.\\
$P_k^i$ and $G_k^i$ represent the coordinates of the i-the joint of the k-the predicted skeleton and the ground truth of that joint, respectively.\\
$\delta(.)=\begin{cases}
  1 & \text{if $\left|\left|P_k^i-G_k^i\right|\right|_2^2\leq t_j$ is True} \\
  0 & \text{otherwise}
\end{cases}$\\

\textbf{\gls{ap}} is calculated using \gls{mpjpe} (see Eq. \ref{eq:mpjpe}) as the thresholding metric between the ground truth and the predicted keypoints, (see Eq. \ref{eq:AP}).
For example, AP@50 refers to the \gls{ap} calculated considering a pose correctly estimated if the $\text{\gls{mpjpe}} < 50mm$.
\begin{equation}\label{eq:AP}
    AP@k = \frac{1}{P}\sum_{p=1}^P\frac{TP_{p}}{TP_{p} + FP_{p}}
\end{equation}
TP corresponds to True Positive.\\
FP corresponds to False Positive.\\
P is the number of people in the scene.\\
k is the threshold in $mm$.\\ 

\textbf{\gls{mAP}} is the average value of all \gls{ap} over all considered thresholds, (see Eq. \ref{eq:mAP}).
\begin{equation}\label{eq:mAP}
    mAP = \frac{1}{T}\sum_{k=1}^T AP@k
\end{equation}
T is the number of considered thresholds.\\

\textbf{\gls{auc}} integrates the curve that evaluates the model performance across all \gls{pck} thresholds.\\

\textbf{\gls{mpjae}}
measures the average, across all angles, of the absolute difference in degrees between the actual joint angles and the estimates, (see Eq. \ref{eq:mjae}).
\begin{equation}\label{eq:mjae}
    \text{MPJAE} = \frac{1}{N}\sum_{i=1}^N \left|\left(x_i-x_i^{'}\right)\mod\pm180\text{º}\right|
\end{equation}
$N$ is the total number of joints.\\
$P_i$ is the estimated pose vector.\\
$P_i^{'}$ is the ground-truth pose vector.\\
$\mod$ is the modulus operator, the term $\text{mod}\pm180$º applies to angles brings them into the range of $[-180\text{º},+180\text{º}]$.\\

\textbf{\gls{mrpe}} consists of the mean Euclidean distance between the predicted root localization and the ground-truth root localization, (see Eq. \ref{eq:mrpe}).
\begin{equation}\label{eq:mrpe}
    \text{MRPE} = \frac{1}{N}\sum_{i=1}^N \left|\left|R^{(i) '}-R^{(i)}\right|\right|_2
\end{equation}
$R^{(i)'}$ is the coordinates of the predicted root.\\
$R^{(i)}$ is the coordinates of the ground-truth root.\\

%% file: sections/Multi-view.tex
\section{Multi-view approaches}\label{sec:multiview}
The availability of multiple views, on the one hand, provides more information for constructing the 3D pose, on the other hand, introduces a new problem, which is how to associate these multiple views. \cite{Integrating_Multiple_Uncalibrated_Views} 

Most of the earlier methodologies were developed to address this task in single-person scenarios, nonetheless, the majority of the more recent approaches focus on multi-person settings. Therefore, the existing methods can be divided into single-person (Section \ref{sec:single}) or multi-person approaches (Section \ref{sec:multiple}), according to the number of people in the scene. 

\begin{figure*}[!htb]
    \centering
    \includegraphics[width=\linewidth]{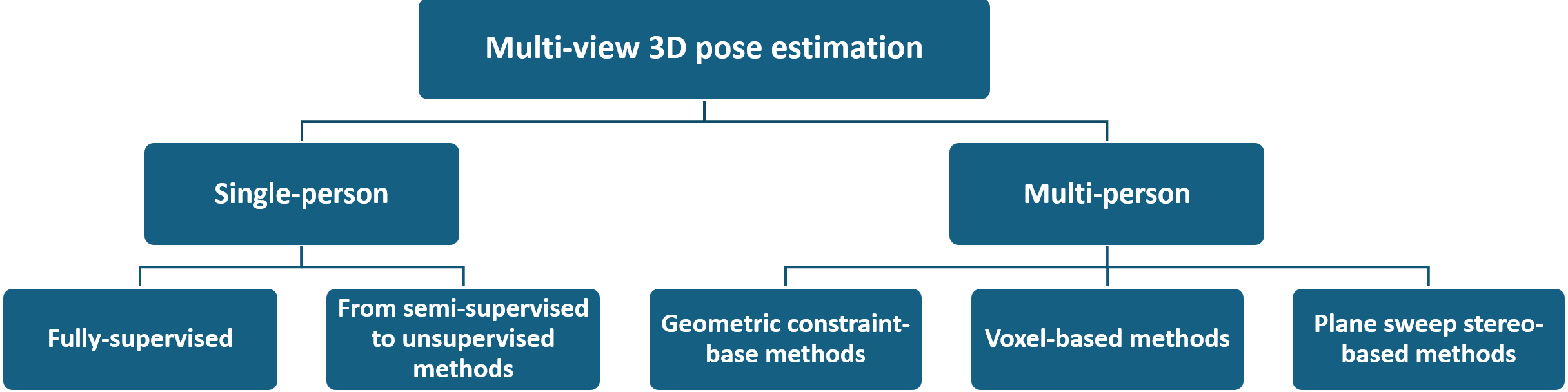}
    \caption{Proposed taxonomy for multi-view 3D pose estimation}
    \label{fig:taxonomy}
\end{figure*}

The proposed taxonomy, see Figure \ref{fig:taxonomy}, is as follows: single-person approaches were divided into fully-supervised, and methods from semi-supervised to unsupervised. The majority of the existing approaches use geometric relationships to infer the 3D pose, nonetheless, some researchers have opted for fully-supervised methods, as others have chosen methods with lower levels of supervision to enhance the generalisation capabilities of their methods and possibly improve their models' performances. Regarding multi-person approaches, those were categorised taking into consideration the type of approach to use to infer the 3D pose from the images. Thus, the methods were categorised into geometric constraint-based if their implementation was based on geometric relations between the camera perspectives or the human body structure. Then, the methodologies under the category of voxel-based methods are approaches which don't rely on 2D pose estimates to determine the 3D pose and use the voxel features to infer the 3D pose. Finally, plane sweep stereo-based methods are approaches that have used the plane sweep stereo network to incorporate depth information into the estimation of the 2D poses to minimise the impact of mismatches.

\subsection{Single-person approaches}\label{sec:single}
In this section, it's presented methods developed to retrieve the 3D pose of scenarios with just one individual. These methods are divided into two categories: Fully-supervised methods (Section \ref{sec:fullysup}), and From Semi-supervi-sed to Unsupervised methods (Section \ref{sec:less_sup}) which include semi-supervised, weakly-supervised, self-supervised and unsupervised methods. This classification is based on the quantity of labelled data used during training. Therefore, models utilizing fully annotated datasets in their training process are designated as fully-supervised methods, while those utilising little to no annotated datasets are designated under the second category. The difference between semi-supervised, weakly-supervised, self-supervised and unsupervised methods is mainly due to the amount of labelled data used for model training. Therefore, semi-supervised methods are trained with more labelled data than weakly-supervised, which in turn are trained with more annotated data than unsupervised methods, which are trained without any labelled data. Self-supervised approaches are a sub-group of unsupervised methods in which implicit labels are generated to help understand the unstructured data. Unsupervised methods, generally, learn to analyse and understand the data structure without relying on labelled data, clustering it into different groups based on the observed characteristics and relationships between the different data examples.

\subsubsection{Fully-supervised methods}\label{sec:fullysup}
Sigal et al. \cite{humaneva} introduced the HumanEva datasets to address the lack of datasets for multi-view 3D pose estimation and provide a benchmark so that developed methods could be compared fairly. Thus, to provide intuition on how to work with their dataset, a baseline model was proposed which consisted of a Bayesian filtering method whose parameters were optimised using Sequential Importance Resampling and \gls{apf}. 

To try to create a method which provides a fast inference while still giving a good performance, Wang and Chung \cite{Integrating_Multiple_Uncalibrated_Views} employed a bottom-up approach by first, determining all possible body part candidates for each view. Then, used a linear-combination expression to determine which detected parts had correspondence across all views, thus decreasing the number of erroneous candidates. Finally, a belief propagation process is applied to group all the parts into a 3D pose.

Mehrizi et al. \cite{mehrizi2018markerfree} followed a two-stage approach. Thus, the researchers proposed a \gls{dnn} composed of a perceptron network to determine the 2D poses and the hierarchical texture information for each view, and by a half-hourglass with skip connections which uses the mentioned information to infer the 3D pose. Showing the importance of sharing the texture information to improve the estimation of the 3D pose.

Amin et al. \cite{Amin2013MultiviewPS} for 3D single-person estimation, used the \gls{ps} model to estimate the 2D poses and then, triangulation to recover the 3D pose. Although the employed method has demonstrated good performance on the HumanEva-I dataset, it's dependent on the camera setup to learn the pairwise appearance terms. 

As for Remelli et al. \cite{remelli2020lightweight}, they presented an efficient method of reconstructing the pose without substantially increasing the computational cost. To do so, the authors introduced a faster version of the \gls{dlt} method to lift the predicted 2D skeletons to 3D poses. The 2D poses are generated by exploring 3D geometry and camera projections.

Solichah et al. \cite{Solichah2020OpenPose} applied the OpenPose \cite{openpose} algorithm, which allows the implementation of a real-time keypoint detector for multi-persons on single images, to forecast the 2D poses. Then, used triangulation to achieve the final 3D pose. Although the method produced satisfactory results, it requires a proper calibration of the camera settings. The approach proposed by Kadkhodamohammadi and Padoy \cite{kadkhodamohammadiandPadoy2019generalizable} uses the following pipeline: estimation of the 2D poses using a single-view detector, match of the 2D poses across views using epipolar geometry and regress the 3D pose using a multilayer neural network composed of multiple stages. The authors have shown that the inclusion of distinct perspectives enhances the model's performance. Nonetheless, one major drawback of their method is the high reliance on the camera calibration settings and the features of the single-view pose detector.

The study conducted by \citet{WAN2023_Viewconsistency} seeks to improve the correlation of 2D keypoints across multiple views by employing a Multi-view Fusion module and to refine the 3D pose estimation through the utilisation of Holistic Triangulation with anatomy constraints. The Multi-view Fusion module integrates pseudo-heatmaps from various views, created based on detected keypoints in the reference view, with the initial heatmap to produce a more precise heatmap. Holistic triangulation with anatomy constraints is designed to simultaneously correlate all views through the integration of a re-projection term based on multi-view geometric constraints and a \gls{pca} reconstruction term to enforce anatomical consistency. Despite its rather good performance, the model encounters challenges in reconstructing poses for unseen movements and lacks adaptability to systems featuring a different number of cameras than those used during training.

As evidenced by previously revised methods, numerous methodologies rely on a fixed camera setup and are unable to generalise to new perspectives. To overcome this issue, Bartol et al. \cite{Bartol2022Triangulation} suggested a triangulation method based on stochasticity. Which after generating the 2D poses for each view, randomly chooses a subgroup of views and combines the corresponding 2D poses through triangulation to produce the 3D pose. This second step is repeated several times to create various hypotheses, to which a score is assigned to allow the computation of the weighted average of all hypotheses and consequently, choose the most appropriate hypothesis. Jiang et al. \cite{ProbabilisticTriangulation_Jiang} have proposed a method, called Probabilistic Triangulation, that approximates the camera pose distribution. The parameters of this distribution are initialised using 2D heatmaps estimated from the input RGB images and are then continuously updated using Monte Carlo sampling. This approach reduces the reliance on calibration parameters, making it possible to estimate the 3D pose in uncalibrated settings. Nonetheless, it still requires the model to be trained with calibrated data.

Nakatsuka and Komorita \cite{NakatsukaKomorita2022FewViews} developed a method to obtain robust 3D pose estimations in demanding environments with limited views and low-resolution data. In this type of environment, the occurrence of occlusions and the lack of visual information are very frequent, damaging the prediction of the pose, to overcome this challenge some researchers have used temporal consistency. Nonetheless, Nakatsuka and Komorita argue that imposing temporal consistency may suppress abrupt and tiny changes which might hinder the overall model's performance. Thereby, they proposed a method composed of two components: one for regressing the 3D pose using images as input, and the other for refining the previous 3D pose estimation, utilising a gated temporal convolution network to correct just the keypoints predicted with poor confidence. The authors claim that their approach could be further optimised by employing a lightweight 3D pose estimator.

Xia and Zhang introduced VitPose \cite{XiaZhang2022VitPose}, a model composed of a CNN backbone used to capture the low-level features; the encoder part of the Vision Transformer to capture the long-range relationship between the human body joints in one perspective and their association with the joints in another perspective; and the Simple Feature Fusion Network, which allows to weightily fuse views to prevent views that give poor pose predictions from harming model performance. Hence, this method provided a more robust fusion step and also, showed that the exploration of long-distance relationships enables more exact 2D pose estimations. \citet{FusionFormer_2024} also explored the fusion of features. However, the researchers focus on fusing features from multiple frames with those from various perspectives. To achieve this, a transformer encoder is used to create a global feature that combines all the features from different views and frames. Then, a transformer decoder is used to fuse the global features with the features specific from each view, yielding a more meaningful set of features to estimate the 3D pose. The employment of this fusion scheme proved to be effective in mitigating the depth uncertainty impact, in addition to demonstrating the potential of methodologies that do not require prior knowledge of camera parameters.

The computational load imposed by \gls{cnn}-based models may limit their application in real-world scenarios. To surpass this problem, Hwang et al. \cite{Hwang2023distributedasynchronous} designed a system consisting of several edge devices and a central server. In which each edge device was responsible for estimating the 2D pose of the received image using a \gls{cnn} and then, sending it along with the respective timestamp to the central server. The central server, in turn, triangulates using \gls{dlt} the received 2D poses that were detected at roughly the same time to produce the 3D pose. Hence, this system distributes the computational burden of the CNN among edge devices and also reduces data traffic by transmitting only the 2D poses and timestamps rather than RGB images, making it more suitable for real-time applications in the real-world.

%\subsubsection{Methods under different supervision levels}
%\subsubsection{Semi-supervised, Weakly-supervised and Unsupervised methods}
\subsubsection{From Semi-supervised to Unsupervised methods}\label{sec:less_sup}
The scarcity of labelled datasets significantly restricts the models' outcomes. Therefore, unsupervised, weakly-supervised, semi-supervised and self-supervised methods have been proposed to surpass the dependence on 3D labelled datasets. 

An example is the work of Rhodin et al. \cite{rhodin2018unsupervised} which employed a semi-supervised method capable of learning from unlabelled images, a geometry-aware representation of the human body. Then, used some supervision to learn the mapping from the 3D geometry representation to the 3D pose. Inspired by this methodology, Kundu et al. \cite{kundu2020kinematicstructurepreserved} introduced an unsupervised method that uses a geometry-aware bottleneck to generate a comprehensible latent space for representing the 3D poses. Rochette et al. \cite{rochette2019weaklysupervised}, on the other hand, explored a weakly-supervised method which uses a multi-view consistency loss and a re-projection consistency loss to determine the 3D pose using only one image. 
Similarly, Ma et al. \cite{Ma2023selfsupervised} also proposed a self-supervised approach in which the loss function leverages information from more than one view to completely disentangle the camera perspective from the 3D human body skeleton and consequently, surpass the projection ambiguity issue observed in a monocular setting. During their research, the authors concluded that the use of more views throughout the training process would further boost the performance of their network. 

In the research work carried out by \citet{TriangulationResidualLoss}, a new loss function, Triangulation Residual Loss, has been proposed, whose objective is to minimise the total distances between view rays and 3D pose estimation retrieved through triangulation. This novel loss function aims to enforce multi-view geometric consistency, facilitating the efficient self-supervised training of the model. As for Jenni and Favaro \cite{JenniFavaro2021selfsupervised}, they proposed a pre-training technique using self-supervised learning, to determine the synchronisation between two images and whether one image is inverted horizontally in relation to the other, with the objective of learning a meaningful representation suitable to be used by a network to estimate the 3D pose without relying on knowing any camera parameter. Also, the authors suggest removing the background for static images to prevent irrelevant features from being learnt during the self-supervised learning task. In addition, Liu et al. \cite{Liu2021dualview} also, introduced a method which relies only on two views to reconstruct the pose, without requiring to know any camera configurations. So, their method after extracting the 2D poses of each image, used a self-supervised 3D regression network able to produce virtual views using orthogonal projections of the human body, allowing the model to thoroughly understand the human body spatial structure and the transformations necessary to project one perspective onto another. 
Concerning the work of \citet{YIN2024_infantpose}, a self-supervised method incorporating temporal convolution blocks and spatio-temporal attention mechanisms has been investigated. The methodology employed involved the exploration of re-projection and view consistency methods, which enabled an accurate retrieval of the 3D pose of infants without the need for 3D labels or definition of camera calibration parameters.

\begin{figure}[!htb]
    \centering%0.95
    \includegraphics[width=\linewidth]{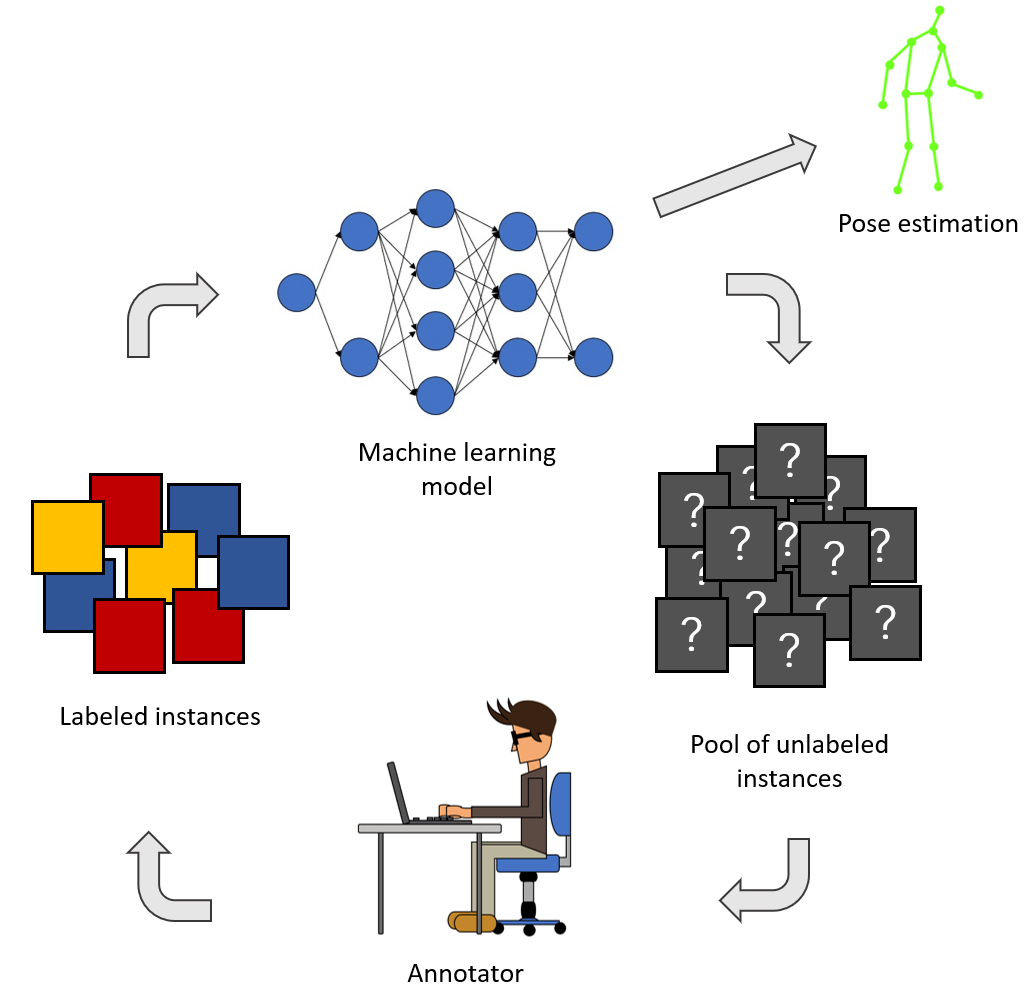}
    \caption{Active learning process. In this process, the model interactively selects some instances and asks the human annotator to label them. Then, the model is re-trained using these new labelled instances until it reaches a defined performance threshold or the budget limit.}
    \label{fig:active_learning}
\end{figure}

%\subsubsection{Self-supervised methods}
In contrast with the previous works, Feng et al. \cite{dataanotation_2023_WACV} proposed a method to enhance the data labelling process instead of trying to learn from the available unlabelled data or producing new synthetic data. Feng et al. have used an Active learning-based methodology (see Figure \ref{fig:active_learning}) to efficiently annotate the data without considerably increasing the overall computational cost. Moreover, the researchers have shown that this strategy employed whether alone or in conjunction with self-training can boost the labelling of the data and consequently, enhance the overall 3D pose estimation outcomes.

Table \ref{tab:h36m_results} shows the results reported in the revised works that have used Human3.6M dataset to evaluate the model's performance. Appendix \ref{app:singleperson} shows the outcomes obtained using single-person techniques on other datasets.

\begin{table}[!htb]
\centering
\scriptsize
\caption{Table summarising the results reported in the revised publications for the Human3.6M \cite{h36m_pami}}
\begin{tabular}{m{1.2cm}m{3.1cm}m{0.5cm}M{1.65cm}}
%\begin{tabular}{m{2.6cm}m{8.2cm}m{0.8cm}M{2.2cm}}
\toprule
\textbf{Type of method} & \textbf{Paper} & \textbf{Year} & \textbf{MPJPE (mm) $\downarrow$} \\ \midrule
\multirow{14}{\hsize}{\textbf{Fully-supervised}} & \citet{joint_multiview2018} & 2018 & 99.70 \\ %\cmidrule{2-4}
 & \citet{kadkhodamohammadiandPadoy2019generalizable} & 2019 & 57.90 \\ %\cmidrule{2-4}
 & \citet{remelli2020lightweight} & 2020 & 30.20 \\ %\cmidrule{2-4}
 & \citet{remelli2020lightweight} \newline(trained with extra data from the MPII Human Pose dataset~\cite{andriluka_mpii}) & 2020 & 21.00 \\ %\cmidrule{2-4}
 & \citet{Reddy2021tessetrack} & 2021 & 18.70 \\ %\cmidrule{2-4}
 & \citet{WANG2022SmartVPoseNet} & 2022 & 67.20 \\ %\cmidrule{2-4}
 & \citet{Wang2022LHPEnetsAL} & 2022 & 31.17 \\ %\cmidrule{2-4}
 & \citet{Bartol2022Triangulation} & 2022 & 29.10 \\ %\cmidrule{2-4}
 & \citet{XiaZhang2022VitPose} & 2022 & 17.00 \\ %\cmidrule{2-4}
 & \citet{ProbabilisticTriangulation_Jiang} & 2023 & 27.80 \\ %\cmidrule{2-4}
 & \citet{WAN2023_Viewconsistency} & 2023 & 21.10 \\ %\cmidrule{2-4}
 & \citet{FusionFormer_2024} & 2024 & \textbf{15.10} \\ %\cmidrule{2-4}
 & \citet{FusionFormer_2024} using the 2D ground-truth poses & 2024 & 7.90 \\ \midrule

%\multirow{7.5}{\hsize}{\textbf{Semi-supervised to \\ Unsupervised}}
\textbf{Semi-supervised} & \citet{rhodin2018unsupervised} & 2018 & 131.70 \\ \midrule %%\cmidrule{2-4}
\multirow{4}{\hsize}{\textbf{Self-supervised}} & \citet{JenniFavaro2021selfsupervised} & 2021 & 64.90 \\ %\cmidrule{2-4}
 & \citet{Liu2021dualview} & 2021 & 22.50 \\ %\cmidrule{2-4}
 & \citet{Ma2023selfsupervised} & 2023 & 75.10 \\ %\cmidrule{2-4}
 & \citet{TriangulationResidualLoss} & 2023 & 25.80 \\ \midrule
\textbf{Unsupervised} & \citet{kundu2020kinematicstructurepreserved} & 2020 & 56.10 \\ %\midrule %\cmidrule{2-4}
\bottomrule
\end{tabular}\label{tab:h36m_results}
\end{table}

\subsection{Multi-person approaches}\label{sec:multiple}
This section presents the solutions developed to determine 3D poses in multi-person scenarios. As such, this section will be divided according to the proposed taxonomy, into geometric constraint-based methods (Section \ref{sec:geo}), voxel-based methods (Section \ref{sec:voxel}) and plane sweep stereo-based methods (Section \ref{sec:pss}).

The approaches have different characteristics, thus, geometric constraint-based methods leverage from using constraints regarding the structure of the human body, epipolar geometry, geometric relationships between views, and similar conditions. Plane sweep stereo-based methods also use geometric constraints and try to relate 2D poses from various perspectives to create the final 3D pose, nonetheless, they rely on depth information calculated using the plane sweep stereo-based model. The voxel-based methods have a more distinct approach, in which the 3D pose is regressed based on a set of features obtained in the 3D space and not on the 2D space which is the most common approach by the other two mentioned categories, preventing these models from being subjected to the performance of a 2D pose estimator and matching processes. Nonetheless, usually are more computationally expensive and slower.

\subsubsection{Geometric constraint-based methods}\label{sec:geo}
Belagiannis et al. \cite{Belagiannis2014} extended the Amin et al. \cite{Amin2013MultiviewPS} approach for multi-person pose estimation, using triangulation of the detected 2D body joints in the various views to create the 3D discrete state space with all body parts hypotheses for all humans in the scene. Then, the \gls{3dps} model was used to infer the 3D human poses from the 3D body part hypotheses in the reduced state space. Later, in \cite{Belagiannis2015_temporallyconsistent}, the researchers introduced temporal consistency to their previous approach to account for estimating human body pose over time. The researchers argue that temporal consistency is highly important for estimating multiple human poses, where the trajectory of each individual is closely related to their body posture, as it aids in penalising false positive candidates. 
\citet{Chen2024321_MovementFunction} aimed to develop a system that can evaluate the similarity between the movements of an ordinary individual and those of someone with a motor dysfunction. To achieve this, the authors integrated temporal information into \gls{3dps} to enhance the 3D pose estimates. 

In \cite{belagiannis20163Dpami}, the authors based their work on their previous work \cite{Belagiannis2014}. Nonetheless, they used different body parametrisations by exploring the use of a \gls{3dps} with several potential functions and also, to learn the model's parameters was used a structured \gls{svm} to adequately weight the different potential functions. 
Zhu et al. \cite{articulated_human_motion} also, explored the \gls{ps} model, but for human motion tracking. To accomplish that, the researchers first, subtracted the background, then, used \gls{fmp}, which is based on \gls{ps}, for foreground learning to detect the human body parts and finally, to track the body parts, it was utilised \gls{apf}. 
The latest work by Belagiannis et al. \cite{belagiannis20163Dpami}, as mentioned, relied on a structured \gls{svm} to optimise the model parameters. In contrast, Schwarcz and Pollard \cite{3DHPE_deepMV_2Dpose}, to simplify the hyperparameters optimisation and improve the quality of the poses estimated with triangulation, used, to determine the 3D pose, a \gls{crf} as a factor graph, the 3D limb locations as variables, and the limb position priors, collision terms and temporal smoothing on those joints as factors. Furthermore, for determining the 2D poses for each view, Schwarcz and Pollard took advantage of the OpenPose library \cite{openpose}.
On the other hand, Ershadi et al. \cite{ErshadiNasab2018MultipleH3} employed a bottom-up approach starting by leveraging from the DeeperCut model \cite{DeeperCut} to detect the 2D body parts. Next, a 3D search space is constructed with all detected joints and a \gls{gmm} is applied to cluster the points in order to obtain the number of people in the scene. To infer the 3D pose, a fully connected pairwise \gls{crf} with its pairwise terms defined in the 3D space and a loopy belief propagation are utilized. 
To alleviate the computational cost of deep neural networks developed to estimate the 3D pose, Wang et al. \cite{Wang2022LHPEnetsAL} proposed LHPE-nets. The LHPE-nets incorporate a low-span network to increase the speed of the start of the training process for the 2D pose prediction as well as a residual deep neural network trained on low-resolution data which showed to enhance network scalability and performance when compared to a ResNet-34.

To deal with the inefficiency of \gls{ps}-based models, Dong et al. \cite{dong2019fast} proposed a multi-way matching algorithm which leverages epipolar geometry, appearance similarity and cycle consistency to reduce the state space and eliminate false detections, resulting in a more efficient matching of 2D poses between views. Later, in \cite{fast_and_robust2022}, the researchers extend their approach to also, track the poses by introducing temporal tracking and Riemannian Extended Kalman filtering. However, their methods present some limitations, like a 2D pose can be deemed an outlier if it only appears in one view, or if there are fast movements, the tracking algorithm is unreliable. 
Thus, their method could not be used in applications requiring quick movement monitoring, such as sports tracking. Bridgeman et al. \cite{bridgeman2019sports}, to improve the speed of the 2D pose association across views, employed a fast greedy algorithm based only on geometric relations and then, used triangulation to generate the 3D skeletons. Although, the presented method is faster than \gls{ps}-based methods, in scenarios with a small number of cameras the use of \gls{ps} for joint estimation has been proven to be more precise than triangulation \cite{fast_and_robust2022,bridgeman2019sports}. Demonstrating the existing trade-off between speed and accuracy \cite{fast_and_robust2022}.

Xu et al. \cite{human_robot_interaction2020Xu} focus on the application of human-robot interaction. So, for that the authors need to conjugate two tasks: estimation of the 3D human skeleton and controlling of the robot movement. Focusing on the pose estimation part, Xu et al. used a 2D pose detector to determine the pose for each view, next, used a greedy algorithm to associate the 2D poses and finally, computed the 3D pose based on paired 2D poses. To enhance efficiency and avoid pose mispairing, the authors implemented a redundancy screening phase to remove redundant pairings, as well as iteration reversal which is used to double-check the validity of the current state; if the state fails this check, the iteration returns to its backup state.

To tackle the issue of swiftly and accurately predicting the 3D pose in crowded scenarios, Chen et al. \cite{crowded2020Chen} used the Jonker-Volgenant algorithm to first, associate the foot joints across the various views, and then, used that information and the human body kinematics to find the matches for the rest of the joints. Lastly, to regress the 3D pose, the researchers refined the triangulation method by introducing \gls{map} optimisation that takes into account the uncertainties of the 2D estimations and imposes the average lengths of the 3D bones.

Huang et al. \cite{end2endDynamic2020} proposed a method which allows a back-propagation of the gradients from the 3D estimation step to the 2D pose detection, enabling end-to-end training of the model. Also, another contribution is the proposed bottom-up dynamic matching algorithm that constructs 3D pose sub-spaces by projecting each pair of 2D poses using triangulation into 3D poses. The matching 3D poses are then identified using a distance-based clustering approach.

As for Chu et al. \cite{Chu_2021_partaware}, the researchers introduced a method to lessen the computational burden imposed by the matching process. The approach consisted of the use of temporal consistency and part-aware measurement to be able to exploit previously obtained 3D poses to discover better 2D-3D correspondence across perspectives. Additionally, to remove possible outliers or noisy data, a joint filter is applied, improving the robustness of the 3D reconstruction. 
On the other hand, Dehaeck et al. \cite{Dehaeck2022Geometric} showed that it's possible to obtain a reliable cross-view matching algorithm by relying only on geometric constraints, like the Sampson error which allows to compute the distance between predicted points from two distinct views. Furthermore, the authors claim that solving the matching problem between views requires at least three perspectives, however, if only two of them are free from occlusions, the trustworthiness of the matches drops.
\citet{Xu2022_MultiviewMultipersonUncalibrated} suggested employing multi-view geometry to reduce the solution space for cross-view matching. Thus, the authors established several requirements: each individual must be within the field of view of at least two cameras, no person can have more matches than the number of available cameras, and individuals from the same view cannot be matched. Then, to enhance the association of the 2D poses across the several views, a self-validation method is used, exploiting the correspondences with higher quality from all camera pairs. Lastly, the 3D poses from different pairs of cameras are triangulated to generate the final 3D pose, which is then refined via bundle adjustment. This methodology culminates in a method robust to distinct camera settings without requiring to know any calibration parameters. 
Ershadi-Nasab et al. \cite{Ershadi-Nasab2021Adversariallearning} adopted an adversarial learning methodology aiming to estimate the 3D pose without requiring any camera calibration. The model is composed of a generator which estimates the 3D pose and attempts to create estimates that the discriminator cannot differentiate from the ground-truth poses. Thus, the generator calculates the 2D Euclidean distance and 2D angular difference matrices before mapping each of them to the equivalent 3D version which are subsequently utilized to generate the 3D pose. Also, for the generator to become more robust to occlusions and improve the 3D pose estimation, a Procrustes analysis was used to allow the association of the poses over several viewpoints. The discriminator compares the 3D Euclidean distance and 3D angular difference matrices of the generated pose with the ground-truth and determines if they are identical. 

Other works have explored techniques to select the best views regarding occlusions to more accurately estimate the 3D pose \cite{joint_multiview2018, Fan2021Multiagent, WANG2022SmartVPoseNet}. Fan et al. \cite{Fan2021Multiagent} developed MAO-Pose, a self-supervised method capable of managing camera positions. Therefore, the various cameras are encouraged to choose the best perspective for the visibility of the joints and to diversify their choices of viewpoints in comparison to the perspectives chosen by the other cameras, in order to promote a broad range of views. Furthermore, a communication system called Consensus is introduced to allow the cameras to exchange their position information and assist the next camera in optimising its position plan by knowing how the other cameras are going to be placed. Wang et al. \cite{WANG2022SmartVPoseNet}, on the other hand, introduced Smart-VPoseNet, a model that relies just on one view to reconstruct the 3D pose. However, the used perspective is chosen dynamically from all available viewpoints in a multi-view system. The choosing criteria that Smart-VPoseNet follows are the number of visible joints, the degree of stretch of the human body and the level of affinity between the perspective and the model. These three criteria can be used individually or in conjunction depending on the characteristics of the dataset. Furthermore, the authors claim that their methodology provides a basis for a possible two-view fusion system.

\subsubsection{Voxel-based approaches}\label{sec:voxel}
\begin{figure*}[!htb]
    \centering%0.95
    \includegraphics[width=\linewidth]{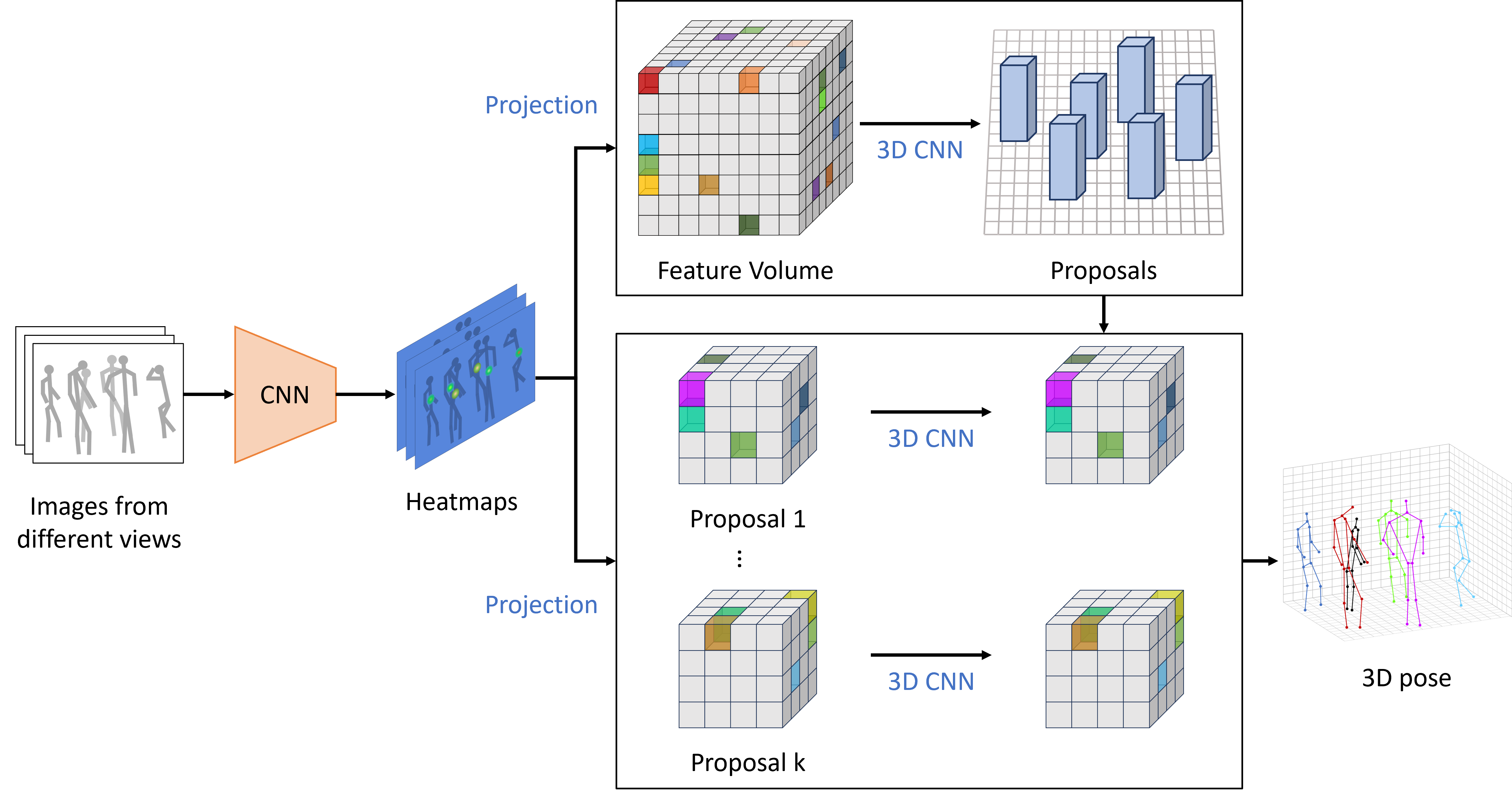}
    \caption{Overview of the VoxelPose architecture (image adapted from \cite{tu2020voxelpose})}
    \label{fig:voxelpose}
\end{figure*}

In \cite{tu2020voxelpose}, Tu et al. introduced VoxelPose (see Figure \ref{fig:voxelpose}), a method that completely avoids erroneous 2D pose associations of different views, by directly working on the 3D space. Thus, the model receives as input the images from the various perspectives and generates the corresponding 2D heatmaps. Next, these heatmaps are projected into the 3D space to form a 3D feature volume which serves as input for the \gls{cpn} that will output a 3D cuboid proposal for each person in the scene. Finally, finer-grained cuboids are created, to enable a more accurate estimation of the pose, and fed to the \gls{prn} to obtain the 3D poses. 

The robustness of the method against occlusions and the superior performance compared to previous approaches has attracted the community to explore this methodology. Thus, various subsequent works are based on this architecture, such as \cite{Deng2023_UnsupervisedDropout,ye2022fastervoxelpose, Zhuang20231763_FasterVoxelPose+, zhang2021voxeltrack, Reddy2021tessetrack, Chen2023VTP, Zhu2023Crowded}. 

\citet{Deng2023_UnsupervisedDropout} in order to make the VoxelPose model robust to domain shift have proposed to include a domain adaptation component, a dropout mechanism and a transferable parameter learning. These new additions will enable the learning of representative features across different domains, while, also, mitigating the adverse consequences of learning domain-specific information. Nonetheless, although it improves cross-domain efficiency and reduces the need for manually labelled data, the training overhead is quite high, restricting its applicability to settings requiring highly demanding processing processes, and this becomes even more pronounced as the number of perspectives increases. 

On the other hand, Faster VoxelPose \cite{ye2022fastervoxelpose} focused on improving the VoxelPose's speed by avoiding the use of 3D \gls{cnn}. In order to do so, the feature volume belonging to each individual had to be identified, and then, re-projected to 2D space independently. However, the method is not robust to the decrease in the number of views, leading to a significant performance drop. Thus, to address that \citet{Zhuang20231763_FasterVoxelPose+} proposed FasterVoxelPose+, introducing two changes to the Faster VoxelPose model: a \gls{dpd} and an \gls{edn}. \gls{dpd} is a projection technique that adds extra depth information to the projection of the 2D pose heatmaps into the 3D voxel features. While \gls{edn} consists of a 2D \gls{cnn} able to combine multi-scale information, allowing 2D re-projected voxel features to be processed in different phases and consequently, reaching more accurate 3D bounding boxes estimates and 2D poses. 
In the case of VoxelTrack \cite{zhang2021voxeltrack}, the authors expanded on the VoxelPose approach to include tracking 3D postures throughout time. Furthermore, a more robust 2D backbone network and extra 3D heatmap supervision of all joints are added to try to improve the 3D pose estimations and, to enhance inference speed, sparse 3D \gls{cnn}s are exploited. However, when the number of cameras decreases, all of these approaches still suffer a decline in performance, with this effect being less pronounced in FasterVoxelPose+. TesseTrack \cite{Reddy2021tessetrack} is, also, less influenced by this, since it integrates temporal information via the 4D spatio-temporal \gls{cnn}. This makes the model more resilient to occlusions, better at predicting joint locations, and able to cope with appearance ambiguities in just one frame. 

Regarding the work of Zhu et al. \cite{Zhu2023Crowded}, the VoxelPose approach was also, followed to compute for each human proposal, the respective feature volume. Nonetheless, as their objective was to create a reliable method to work in crowded environments, the authors have proposed a three-stage strategy to refine the 3D poses. Thus, the initial step was to generate finer-grained and narrower feature volumes in the region around each human proposal. In the second-stage, the Voxel Hourglass Network used those feature volumes to generate 3D heatmaps and tag-maps. Then, a 3D Associative Embedding was employed to combine the information provided by the heatmaps and tag-maps to produce a coarse 3D pose. In the third-stage, a refinement layer is utilised to refine the 3D pose given by the previous step. The authors demonstrated that the incorporation of 3D tag-maps resulted in the elimination of undesired joints coming from other people in the scene which, consequently, lowered the amount of mismatched joints.

Additionally, inspired by VoxelPose, Wang et al. \cite{wang2021direct} introduced MvP, a transformer-based model for solving the multi-person 3D pose estimation task. This method directly predicted the 3D skeleton by encoding the joints as learnable query embeddings and joining them with the features extracted from the input images. Furthermore, to increase the model's performance, the suggested transformer used a projective attention mechanism based on geometrical information in conjunction with a RayConv operation to precisely fuse the cross-view data for each joint. The authors point out some drawbacks of their approach which are the lack of robustness to generalize for new camera perspectives and a high demand for training data. Another transformer-based approach, based on VoxelPose is \gls{vtp} \cite{Chen2023VTP}. The creation of this model aimed to solve the high computational cost associated with applying self-attention directly to volumetric representations. To overcome this, \gls{ssa} mechanism was exploited as it provides quasi-global local attention. To do this, the input sequence is divided into blocks, then the correlation between them is learnt, and finally, \gls{ssa} uses this information to understand how to reorganise and classify the blocks. The authors have concluded that by applying \gls{ssa} the memory consumption was reduced, allowing to efficiently apply self-attention to the volumetric representations.

\subsubsection{Plane sweep stereo-based models}\label{sec:pss}
\begin{figure*}[!htb]
    \centering
    \includegraphics[width=0.9\linewidth]{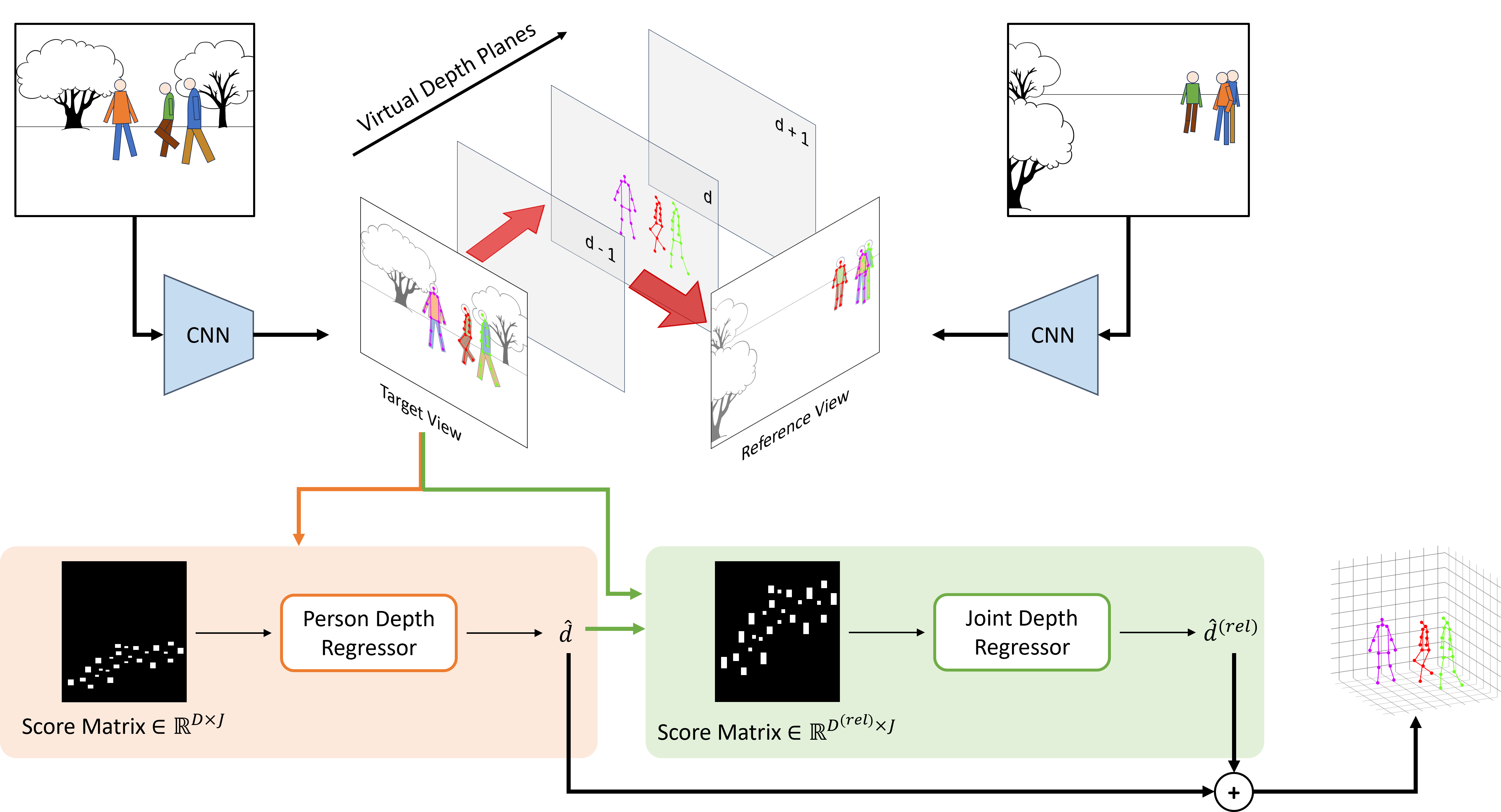}
    \caption{Overview of the plane sweep stereo-based method followed by Lin and Lee (image adapted from \cite{LinandLee_2021_PlaneSweep})}
    \label{fig:planesweepstereo}
\end{figure*}

To create a faster and less computationally expensive algorithm compared to VoxelPose, Lin and Lee \cite{LinandLee_2021_PlaneSweep} proposed a framework based on plane sweep stereo to add person-level and joint-level depth information to the 2D poses generated for each viewpoint, see Figure \ref{fig:planesweepstereo}. Then, the 2D poses and respective depth information are back-projected to create the 3D pose, reducing the computational burden of pose association.  
Extending this approach to work with unlabelled datasets, de França Silva et al. \cite{deFrançaSilva2022unsupervised_planesweep} changed the loss function used on the previously presented plane sweep stereo method to be able to learn the 3D poses in an unsupervised way. Thus, instead of computing the difference between the generated 3D pose and the ground-truth, this novel loss function employs re-projection error between the 2D projection of the 3D pose with the 2D pose estimated for that perspective. The authors besides demonstrating the potential of this approach to learn from data without annotations, also, showed the potential of Adabelief to provide quicker convergences and better performances than the Adam optimiser. Later \citet{WallisdeFrançaSilva2023607_UMVpose++} extended their work by exploring a matching method that uses ground points related to each individual to match the target with the reference view. Additionally, the authors incorporated the smooth $L_1$ loss for computing the re-projection error of the 2D poses. These modifications yielded a significant performance enhancement compared to the previous version, while also showcasing the potential of this method to cope with unlabeled data. However, it is noteworthy that this approach necessitates the knowledge of camera parameters in order to calculate the loss.

Zhou et al. \cite{Zhou2022ER-Net} also, took advantage of the plane sweep stereo method to get the depth score matrices, and subsequently obtain the 3D pose by back-projecting the 2D poses with the corresponding depth information. However, to optimise the prediction of all joints, particularly those most affected by occlusions or other external factors that may impair their detection, the authors employed an attention channel mechanism and an optimal calibration method based on the dependency relationship between the joints and the individual. This strategy promoted channel-wise feature re-calibration by adjusting the weights given to each channel, allowing the model to understand the long dependency relationship between joints and improve the detection of the most difficult joints.

Table \ref{tab:campus_shelf_results} reports the results for all revised works that have used Campus or Shelf dataset to assess the performance of their model. Appendix \ref{app:multiperson} presents the outcomes obtained with multi-person approaches on other datasets.

\begin{table*}[!htb]
\centering
\scriptsize
\caption{Table summarising the results reported in the revised publications for the Campus \cite{Belagiannis2014} and Shelf \cite{Belagiannis2014} datasets.}
\begin{tabular}{M{1.15cm}m{2.6cm}M{0.5cm}M{0.99cm}M{0.99cm}M{0.99cm}M{0.85cm}M{0.99cm}M{0.99cm}M{0.99cm}M{0.85cm}}
\toprule
\multirow{2.5}{\hsize}{\textbf{Type of method}} & \multirow{2}{*}{\textbf{Paper}} & \multirow{2}{*}{\textbf{Year}} & \multicolumn{4}{c}{\textbf{Campus --- PCP (\%) $\uparrow$}} & \multicolumn{4}{c}{\textbf{Shelf --- PCP (\%) $\uparrow$}} \\ \cmidrule(r){4-7} \cmidrule(r){8-11} 
 &  &  & \textbf{Actor 1} & \textbf{Actor 2} & \textbf{Actor 3} & \textbf{Average} & \textbf{Actor 1} & \textbf{Actor 2} & \textbf{Actor 3} & \textbf{Average} \\ \midrule
\multirow{13}{\hsize}{\textbf{Geometric constraint-based}} & \citet{Belagiannis2014} & 2014 & {82.00} & {72.00} & {73.00} & 75.60 & {66.00} & {65.00} & {83.00} & 71.30 \\ %\cmidrule{2-11}
 & \citet{Belagiannis2015_temporallyconsistent} & 2015 & {83.00} & {73.00} & {78.00} & 78.00 & {75.00} & {67.00} & {86.00} & 76.00 \\ %\cmidrule{2-11}
 & \citet{belagiannis20163Dpami} & 2016 & {93.45} & {75.65} & {84.37} & 84.49 & {75.26} & {69.68} & {87.59} & 77.51 \\ %\cmidrule{2-11}
 & \citet{3DHPE_deepMV_2Dpose} & 2018 & {86.55} & {82.54} & {88.14} & 85.15 & {88.34} & {85.26} & {91.30} & 88.92 \\ %\cmidrule{2-11}
 & \citet{ErshadiNasab2018MultipleH3} & 2018 & {94.18} & {92.89} & {84.62} & 90.56 & {93.29} & {75.85} & {94.83} & 87.99 \\ %\cmidrule{2-11}
 & \citet{bridgeman2019sports} & 2019 & {91.84} & {92.71} & {93.16} & 92.57 & {99.68} & {92.79} & {97.72} & 96.73 \\ %\cmidrule{2-11}
 & \citet{dong2019fast} & 2019 & {97.60} & {93.30} & {98.00} & 96.30 & {98.80} & {94.10} & {97.80} & 96.90 \\ %\cmidrule{2-11}
 & \citet{end2endDynamic2020} & 2020 & {97.96} & {94.81} & {97.39} & 96.71 & {98.75} & {96.22} & {97.20} & 97.39 \\ %\cmidrule{2-11}
 & \citet{Chu_2021_partaware} & 2021 & {98.37} & {93.76} & {98.26} & 96.79 & {99.14} & {95.41} & {97.64} & 97.39 \\ %\cmidrule{2-11}
 & \citet{Ershadi-Nasab2021Adversariallearning} & 2021 & {98.41} & {95.12} & {98.82} & 97.45 & {\textbf{99.83}} & {94.91} & {98.33} & 97.69 \\ %\cmidrule{2-11}
 & \citet{fast_and_robust2022} & 2022 & {97.60} & {93.30} & {98.00} & 96.30 & {98.80} & {94.10} & {97.80} & 96.90 \\ %\cmidrule{2-11}
 & \citet{Xu2022_MultiviewMultipersonUncalibrated} & 2022 & {\textbf{99.00}} & {94.70} & {\textbf{99.60}} & \textbf{97.80} & {99.60} & {95.20} & {\textbf{98.50}} & 97.80 \\ %\cmidrule{2-11}
 & \citet{Chen2024321_MovementFunction} & 2024 & {---} & {---} & {---} & --- & {98.60} & {95.80} & {97.90} & 97.40 \\ \midrule
  
\multirow{8}{\hsize}{\textbf{Voxel-based}} & \citet{tu2020voxelpose} & 2020 & {97.60} & {93.80} & {98.80} & 96.70 & {99.30} & {94.10} & {97.60} & 97.00 \\ %\cmidrule{2-11}
 & \citet{wang2021direct} & 2021 & {98.20} & {94.10} & {97.40} & 96.60 & {99.30} & {95.10} & {97.80} & 97.40 \\ %\cmidrule{2-11}
 & \citet{zhang2021voxeltrack} & 2021 & {98.10} & {93.70} & {98.30} & 96.70 & {98.60} & {94.90} & {97.70} & 97.10 \\ %\cmidrule{2-11}
 & \citet{Reddy2021tessetrack} & 2021 & {97.90} & {\textbf{95.20}} & {99.10} & 97.40 & {99.10} & {96.30} & {98.30} & 98.20 \\ %\cmidrule{2-11}
 & \citet{ye2022fastervoxelpose} & 2022 & {96.50} & {94.10} & {97.90} & 96.20 & {99.40} & {96.00} & {97.50} & 97.60 \\ %\cmidrule{2-11}
  & \citet{Chen2023VTP} & 2023 & {97.60} & {93.10} & {98.10} & 96.30 & {99.30} & {95.10} & {97.40} & 97.30 \\ %\cmidrule{2-11}
 & \citet{Zhuang20231763_FasterVoxelPose+} & 2023 & {97.40} & {93.60} & {98.10} & 96.40 & {99.00} & {96.30} & {97.70} & 97.70 \\ %\cmidrule{2-11}
 & \citet{Deng2023_UnsupervisedDropout} & 2023 & {85.10} & {86.30} & {78.40} & 83.20 & {96.50} & {94.10} & {97.70} & 96.10 \\ \midrule
 
\multirow{4}{\hsize}{\textbf{Plane sweep stereo-based}} & \citet{LinandLee_2021_PlaneSweep} & 2021 & {98.40} & {93.70} & {99.00} & 97.00 & {99.30} & {96.50} & {98.00} & 97.90 \\ %\cmidrule{2-11}
 & \citet{deFrançaSilva2022unsupervised_planesweep} & 2022 & {96.90} & {87.80} & {88.90} & 91.20 & {---} & {---} & {---} & --- \\ %\cmidrule{2-11}
 & \citet{Zhou2022ER-Net} & 2022 & {98.60} & {94.00} & {99.20} & 97.30 & {99.40} & {\textbf{97.30}} & {98.20} & \textbf{98.30} \\ %\cmidrule{2-11}
 & \citet{WallisdeFrançaSilva2023607_UMVpose++} & 2023 & {98.40} & {93.40} & {98.60} & 96.80 & {---} & {---} & {---} & --- \\ \bottomrule
\end{tabular}\label{tab:campus_shelf_results}
\end{table*}

%% file: sections/WiFi_RF_signals.tex
\section{Multi-modal approaches}\label{sec:multimodal}
This section presents methods that have used other sensors rather than RGB cameras to estimate the 3D pose.

\subsection{Different type of cameras}
Some researchers have tackled the 3D pose estimation task using types of camera sensors, such as RGB-D cameras \cite{Carraro2019RGBD,Hwang2023rgbd_animation}, \gls{tof} cameras \cite{Hassan2022ToFcameras}, or 360º cameras \cite{Shere2021_360Cameras}  which provide images that differ from the typical RGB images. 
These types of cameras can capture more information which can be advantageous for the estimation of the pose. For example, Shere et al. \cite{Shere2021_360Cameras} created a method to determine the 3D pose using two 360º cameras. The authors state that this type of camera is especially advantageous since it can capture the whole scene without requiring a huge number of cameras while simultaneously providing accurate depth information. Thus, they used temporal information to associate the 2D poses and mitigate inaccuracies. Although their method has been developed to be used with data outputted by 360º cameras, data from other types of cameras may also be used, however, it will require the data to go through a costly re-projection step. 
In the case of Carraro et al. \cite{Carraro2019RGBD}, they have estimated the 3D human pose using an array of \gls{rgb-d} cameras. This allowed them to leverage depth information to compute the 3D pose from the 2D poses generated by using a single-view pose estimation algorithm and the cameras' extrinsic calibration parameters. Additionally, the authors used Kalman Filters to track the poses over time. The developed system thus allows the estimation of multiple people in real-time. Hwang et al. \cite{Hwang2023rgbd_animation} have also, taken advantage of the depth information given by the RGB-D cameras to optimise the process of the cross-view matching of the 2D poses. Then, to be able to deploy this method in real-world, the authors have employed a system similar to \cite{Hwang2023distributedasynchronous}. 

Regarding the work of Hassan et al. \cite{Hassan2022ToFcameras}, the authors have used \gls{tof} cameras which provide infra-red intensity images along with depth information. The use of infra-red intensity images allows the system to be more robust to occlusions, material reflectivity, and lighting conditions in comparison to systems that utilise RGB images. Besides, the authors have proven that the use of intensity images as input is more reliable for estimating the 2D poses and subsequently, projecting them to construct the 3D pose than the depth images. However, in case a joint cannot be determined, the model takes advantage of the depth information to get the position of the missing joint.

\subsection{Wireless sensors}
On the other hand, some researchers have opted to use other types of sensors which can overcome some limitations vision-based methods suffer. Thus, the use of wireless communications to complete this task has been explored, because they aren’t affected by occlusions, privacy concerns or bad illumination conditions which are some challenges vision-based methods have to face \cite{Song2022RF-URL,RF_3Dpose,Xie2024RPM}.

Typically, systems designed to rebuild 3D poses using wireless signals, such as \gls{rf} \cite{Song2022RF-URL,RF_3Dpose,Xie2024RPM} or Wi-Fi \cite{Wang2021CommodityWiFi,Ren2022GoPose}, include a camera that is synchronised with the wireless signals. Nonetheless, the researchers only used the camera data to produce ground-truth 3D poses with which to compare the skeletons generated using wireless signals \cite{Wang2021CommodityWiFi,Ren2022GoPose,Song2022RF-URL,RF_3Dpose,Xie2024RPM}. The typical pipeline for approaches that use these types of signals starts with transforming the input signal into images by, for example, extracting the channel-state information or the angle of arrival spectrums. Then, a method from the computer vision domain is employed to obtain the final 3D pose \cite{Wang2021CommodityWiFi,Ren2022GoPose,Song2022RF-URL,RF_3Dpose,Xie2024RPM}. 

%% file: sections/Discussion.tex
\section{Discussion}\label{sec:discussion}
This section will compare the various approaches for estimating 3D poses in multi-view setups. As a result, it will analyse the benefits and drawbacks of single-person and multi-person methods. %, as well as compare their approaches.

%\subsection{Single-person: Comparing Fully-supervised with Semi-supervised to Unsupervised methods}
\subsection{Single-person: Fully-supervised vs. Semi-supervised to Unsupervised methods}

The majority of the existing methods, for 3D single-person pose estimation, employ the following pipeline: determine the 2D pose for each view based on the corresponding RGB image; associate the 2D poses using geometric relations and then, regress the 3D pose \cite{Amin2013MultiviewPS, Bartol2022Triangulation, FusionFormer_2024, Hwang2023distributedasynchronous, ProbabilisticTriangulation_Jiang, kadkhodamohammadiandPadoy2019generalizable, NakatsukaKomorita2022FewViews, remelli2020lightweight, Solichah2020OpenPose, WAN2023_Viewconsistency, XiaZhang2022VitPose}. 

Most of the methods use well-established 2D pose estimators such as ResNets, \gls{ps} or OpenPose, to determine the 2D poses \cite{remelli2020lightweight, Amin2013MultiviewPS, Solichah2020OpenPose}. Then, match the 2D poses using triangulation or epipolar geometry. However, these matching methods are highly dependent on the number of views and cameras' parameters, setup and calibration. To overcome that, has been suggested the use of probabilistic triangulation to try to determine the camera pose distribution and parameters, however, this method still needed to be trained using data with calibrated camera pose parameters \cite{ProbabilisticTriangulation_Jiang}. The stochastic triangulation method employed by \citet{Bartol2022Triangulation} has shown to surpass both the dependence on camera setup and the number of views by choosing diverse sub-groups of 2D poses determined from different views, combining them, giving a score to each of the generated hypothesis and then, choose the best hypothesis based on the attributed score. Another approach that overcame the reliance on camera parameters was the FusionFormer, which fuses the features from several frames with those from various views \cite{FusionFormer_2024}. This method also mitigated the depth uncertainty by providing the model with more spatial and temporal information in a unified way. In addition, this allowed the authors to achieve state-of-the-art results (see Table \ref{tab:MPI-INF-3DHP} and \ref{tab:totalcapture}), namely, in the Human3.6M dataset (see Table \ref{tab:h36m_results}), in which its performance is 1.9 mm better than the second best performance, \cite{XiaZhang2022VitPose}, which is dependent on the calibration of the camera parameters. Moreover, the performance of the FusionFormer could be further improved with a better 2D pose estimator, as the authors have shown, their model performance using the 2D ground-truth poses achieves a \gls{mpjpe} of 7.90 mm, which would be an improvement of 7.20 mm, corresponding to approximately, 47.7\%. This demonstrates that the 2D pose estimates are one of the biggest limitations of methods that rely on 2D poses as an intermediate step before obtaining the 3D pose. To make more robust 2D pose estimations, \citet{XiaZhang2022VitPose} have exploited the long-distance dependence between joints, reaching a performance of 17.00 mm, corresponding to the second-best result for the Human3.6M dataset. Also, the authors have shown that the fusion process is often affected by less accurate predictions, so, they employ a method which avoids using those predictions.

To make the models more lightweight and faster, authors have suggested the use of \gls{dlt} and Shifted Inverse Iterations to improve the matching speed of triangulation \cite{remelli2020lightweight}, and distribute the work through several edge devices and transmit only the 2D pose estimates instead of the RGB images \cite{Hwang2023distributedasynchronous}. 

Regarding methods with less supervision, \citet{rhodin2018unsupervised} provided, by far, the worst performance of all methods revised for single-person pose estimation with a difference of 116.6 mm to the best method. However, it is to be noted that this method doesn't rely on any 2D or 3D labels to obtain a geometric-aware representation of the human body and only depends on very few labelled data to obtain the 3D pose from that intermediate representation. The approach of \citet{kundu2020kinematicstructurepreserved} didn't rely on any labelled data, nonetheless, it was able to give a performance of 56.10 mm in the Human3.6M dataset which is 57.4\% better than \citet{rhodin2018unsupervised}. This difference is mostly due to the inclusion of kinematic constraints, preventing the model from generating unrealistic pose outputs. Nonetheless, this approach still remains 41 mm away from the best result. 

Other approaches with low levels of supervision have taken advantage of multi-view information and consistency \cite{Ma2023selfsupervised, YIN2024_infantpose, TriangulationResidualLoss, rochette2019weaklysupervised} to help, for example, disentangle the camera views from the 3D poses \cite{Ma2023selfsupervised} or in other cases, avoid the need of camera calibration parameters \cite{YIN2024_infantpose}. In addition, to avoid relying on camera parameters, some researchers have exploited the synchronisation between views \cite{JenniFavaro2021selfsupervised} or synthesis of novel views \cite{Liu2021dualview, rhodin2018unsupervised}. The work of \citet{Liu2021dualview} by generating new virtual perspectives was able to better understand the spatial structure of the human body which allowed them to achieve a performance of 22.50 mm in the Human3.6M dataset, the best performance of a not fully-supervised method. 

A less explored methodology was the use of active learning which helped overcome the lack of annotated datasets in a cost-effective way \cite{dataanotation_2023_WACV}.

Therefore, it can be concluded that fully-supervised methods are, in general, more capable of giving better results than methods with lower levels of supervision. This outcome is to be expected since fully-supervised methods are trained with completely annotated datasets. However, in situations where labelled data is not available, semi-supervised, weakly-supervised, unsupervised, or self-supervised methods are more capable of giving better performances than fully-supervised approaches because they have better generalisation capabilities which allow them to more easily adapt to unseen scenarios. Also, as mentioned the performance of the best self-supervised method is only 7.4 mm worse than the state-of-the-art performance for Human3.6M, see Table \ref{tab:h36m_results}, meaning that with further research it might be possible to create a method able to provide trustworthy 3D pose estimations without needing label data to train, which would potentiate its application in distinct real-world scenarios.

\subsection{Multi-person: Geometric constraint-based vs. Voxel-based vs. Plane Sweep Stereo-based methods}

Similar to single-person approaches, most methods categorised as geometric constraint-based and plane sweep stereo-based depend on 2D pose estimates to obtain the 3D pose. As such, regardless of the method employed for 3D pose estimation, the full potential of these types of models is always constrained by inaccuracies in the estimation of the 2D poses. 

A common approach for geometric-based methods was to use \gls{crf} as a graph to model the human body \cite{Belagiannis2014, Belagiannis2015_temporallyconsistent, belagiannis20163Dpami, Chen2024321_MovementFunction, ErshadiNasab2018MultipleH3, 3DHPE_deepMV_2Dpose}. \citet{Belagiannis2014} have introduced 3D \gls{ps}, a method based on \gls{crf} with multi-view potential functions that allow the model to take into consideration the re-projection errors, the confidence of the 2D body part detections, the visibility of different body parts on the various views, the length of the body parts, as well as human body priors in which a symmetric body parts constraint is employed to avoid collision in the 3D space. This approach allowed the authors to determine the 3D pose either in single-person settings (see Table \ref{tab:KTHMultiviewFootballII} and \ref{tab:HumanEva}) as well as in multi-person scenarios (see Table \ref{tab:campus_shelf_results}) providing a performance of 75.6\% for the Campus dataset and 71.3\% for the Shelf dataset. Then, to further enhance the 3D pose predictions and penalise false detections, the introduction of temporal consistency has been suggested \cite{Belagiannis2015_temporallyconsistent, Chen2024321_MovementFunction, 3DHPE_deepMV_2Dpose}, which has been proven to be able to provide performance enhancements as can be observed by the improvements of 2.4 \gls{pp} in the Campus and 4.7 \gls{pp} in the Shelf dataset comparing the two approaches of Belagiannis et al. \cite{Belagiannis2014, Belagiannis2015_temporallyconsistent}. Later in \cite{belagiannis20163Dpami}, the addition of a \gls{svm} to weight the potential functions has led to even further improvements giving a performance of 84.49\% in the Campus and 77.51\% in the Shelf dataset which is 6.49\gls{pp} and 1.51\gls{pp}, respectively, better than \cite{Belagiannis2015_temporallyconsistent}. 

One of the most critical steps that has been limiting the models' performance is the matching process. \citet{dong2019fast} proposed the use of epipolar geometry, appearance similarity, and cycle consistency to reduce the state space and eliminate false detections. This allowed the authors to obtain a performance of 96.3\% in the Campus dataset and 96.9\% in the Shelf dataset, being 5.74 \gls{pp} and 7.98 \gls{pp} superior compared to the results of \cite{ErshadiNasab2018MultipleH3} in the Campus dataset and \cite{3DHPE_deepMV_2Dpose} in the Shelf dataset which were the best previous approaches based on priors. This approach was later extended, \cite{fast_and_robust2022}, to track people by introducing temporal information and the Riemannian Extended Kalman filtering, however, no further improvements were obtained, in terms of \gls{pcp} results. Other methods have also suggested the use of geometric relations to better match the poses across views \cite{Xu2022_MultiviewMultipersonUncalibrated, Ershadi-Nasab2021Adversariallearning, Dehaeck2022Geometric, end2endDynamic2020}. \citet{Dehaeck2022Geometric} reached the conclusion that is necessary three views and at least two of those views have to be unobstructed for the models to accurately estimate the 3D poses. \citet{Xu2022_MultiviewMultipersonUncalibrated} by imposing that every individual must appear at least in two views, the number of matches for each person can not exceed the number of available views, and that the individuals from the same view cannot be matched, were able to reduce the solution search space and obtain the state-of-the-art performance for the Campus dataset with a \gls{pcp} of 97.80\% and also, for the Shelf dataset, the best result of the approaches categorised as geometric constraint-based models, with a \gls{pcp} result of 97.80\%. Furthermore, temporal consistency, redundancy screening, part-aware measurements, and body kinematics were other techniques also proposed to improve the matching of the poses \cite{Chen2024321_MovementFunction, Chu_2021_partaware}.

Nonetheless, real-world applications require not only accurate 3D pose algorithms but also quick ones, as shown by \citet{fast_and_robust2022}, in which their approach was effective but inadequate for fast motions. \citet{bridgeman2019sports} improve the cross-view matching speed by employing a fast greedy search algorithm to match the 2D poses and triangulation to retrieve the 3D pose. However, even though triangulation provided faster inferences of the 3D pose, methods based on priors are generally more accurate which becomes more pronounced for settings with a smaller number of cameras. Therefore, the use of current approaches in real-world applications is subjected to a trade-off between performance and speed. Other researchers have focused on reducing the computational cost which also improved the training speed by using a light-weight network \cite{Wang2022LHPEnetsAL}. In the case of \cite{Fan2021Multiagent, WANG2022SmartVPoseNet}, the authors have focused on employing methods capable of selecting optimal camera viewpoints that are minimally impacted by occlusions, to accurately estimate the 3D poses. The proposed methods involved either adjusting the camera's position to enhance each camera field of view and obtain more distinct perspectives \cite{Fan2021Multiagent}, or selecting viewpoints that provided better visualisation of the joints, less stretch of the human body and more level of affinity between the view and the model.

Voxel-based methods are also subjected to a trade-off between computational complexity and performance. Although they can deliver highly effective performances, this comes at the expense of high computational complexity. Therefore, voxel-based methods are capable of overcoming the mismatches of 2D poses by directly working on the 3D space. However, to be able to extract the features from 3D volumes, it is required a 3D \gls{cnn} which is computationally inefficient, leading to high computational costs, in terms of complexity and speed \cite{tu2020voxelpose, Deng2023_UnsupervisedDropout, ye2022fastervoxelpose}. To address this issue and make the original VoxelPose model \cite{tu2020voxelpose} more efficient, \citet{ye2022fastervoxelpose} proposed to re-project the 3D features volumes onto the three 2D coordinate planes which will allow the use of 2D CNNs for the estimation of partial 3D poses, that are then fused to obtain the final 3D pose. This approach improved the VoxelPose speed by 10 times without significantly affecting the overall performance, as can be observed in Table \ref{tab:campus_shelf_results}, for the Campus dataset, there was a performance drop of 0.5 \gls{pp}, nonetheless, for the Shelf dataset, Faster VoxelPose \cite{ye2022fastervoxelpose} was able to achieve a \gls{pcp} result 0.6 \gls{pp} superior to VoxelPose \cite{tu2020voxelpose}. Furthermore, VoxelTrack \cite{zhang2021voxeltrack} has proposed to use sparse 3D CNN to improve inference speed. Nonetheless, although both methods could improve the speed of the VoxelPose network, their performance would drastically decline if the number of views was reduced. Thus, \citet{Zhuang20231763_FasterVoxelPose+} has suggested the incorporation of depth information as well as the fusion of multi-scale information which mitigated the performance drop with the decrease in the number of views. On the other hand, TesseTrack \cite{Reddy2021tessetrack} proposed the integration of temporal information into the VoxelPose architecture, which allowed the model to be less affected by occlusions, the decrease in the number of views and appearance ambiguities, consequently, leading to a best 3D pose estimation. This method obtained the best performance of all voxel-based methods with a \gls{pcp} result of 97.40\% for Campus and 98.20\% for Shelf, which means an improvement of 0.7\gls{pp} and 1.2\gls{pp} compared to the original VoxelPose model \cite{tu2020voxelpose}. Nevertheless, TesseTrack is very computationally demanding requiring the authors 8 V100 GPUs with 32GBs of memory to be able to train their model. 
In order to try to improve the computational cost, \citet{Chen2023VTP} have focused on solving the high computational cost associated with self-attention mechanisms applied to volumetric representations, demonstrating that the use of \gls{ssa} allows employing self-attention to volumetric representations without requiring unmanageable amounts of memory.

\citet{wang2021direct} is the only approach, within voxel-based methods, that directly predicts the 3D pose, using only the input image features and joint information. However, this method requires high quantities of data for training and has difficulties in generalising to new views. In terms of performance, it is inferior to VoxelPose \cite{tu2020voxelpose} in the Campus dataset by 0.1 \gls{pp} and 0.4 \gls{pp} superior on the Shelf dataset.

Finally, regarding the Plane sweep stereo-based methods, the approaches were based on the initial work of \citet{LinandLee_2021_PlaneSweep} which added depth information at the joint and person levels to the 2D poses, thus, creating a more accurate, fast and computationally less expensive approach than VoxelPose method \cite{tu2020voxelpose}. As can be observed, in Table \ref{tab:campus_shelf_results}, \citet{LinandLee_2021_PlaneSweep} got a performance 0.3 \gls{pp} and 0.9 \gls{pp} better than VoxelPose \cite{tu2020voxelpose} in Campus and Shelf dataset, respectively. To further optimise the joints' detection and the long dependency relationships between them, \citet{Zhou2022ER-Net} has proposed the use of a channel attention mechanism and optimal calibration method leveraging from the joints-person relationship. This resulted in a performance improvement of 0.3 \gls{pp} and 0.4 \gls{pp} compared to the original approach \cite{LinandLee_2021_PlaneSweep}, and in the state-of-the-art performance for the Shelf dataset with an outcome of 98.30\%.

de França Silva et al. \cite{deFrançaSilva2022unsupervised_planesweep, WallisdeFrançaSilva2023607_UMVpose++} show that is possible to train this approach in an unsupervised manner by changing the loss function to compare the re-projection error of the 2D pose projections from the 3D pose with the 2D poses estimated for that view. This initial approach allows the authors to achieve a performance of 91.2\% on the Campus dataset. Then by improving the matching method to associate the points in the target view with those in the reference, led the model to improve its performance by 5.6 \gls{pp}, being just 0.2 \gls{pp} away from the performance of \citet{LinandLee_2021_PlaneSweep}, which is trained with fully annotated data.

In summary, most approaches take advantage of geometric relations among various views and also, impose constraints associated with the human body structure. Analogous to single-person approaches, the most common pipeline involves having the estimation of the 2D pose as an intermediate step, thereby making the performance of these models in 3D pose estimation conditioned on the quality of the matching processes. To make the models more robust to the association of the poses, it is suggested to reduce the search space by imposing, for example, geometric or visibility constraints. Moreover, the addition of temporal consistency and depth information has also shown promising results in enhancing the models' matching performance.

Voxel-based methods are suggested as a way to completely avoid the mismatch issue, however, they do this at the expense of higher computational time and complexity, which also, heavily limits their employment in real-world settings. As evidenced by \cite{bridgeman2019sports, dong2019fast}, the models need to balance both high accuracy but also, light networks and quick inference times to be useful across numerous real-world applications.

In terms of accuracy, the average performance in both Campus and Shelf datasets (see Table \ref{tab:campus_shelf_results}) is the same for the best model in each of the defined categories. Thus, looking only at \gls{pcp} results, there is no clear favourable category. However, considering also, the computational requirements and speed, the geometric constraint-based methods and plane sweep stereo-based methods are preferable compared to voxel-based methods. Nonetheless, more work is still necessary to find an efficient and effective algorithm to be applied in diverse real-world situations.

\subsection{Opportunities for improvement}
Most methods either for single-person or for multi-person are based on geometric constraints which highly rely on the correct matching of 2D pose estimations. Thus, the 3D predictions are always limited to the accuracy of the 2D pose estimators. Moreover, even though, more recently, some researchers have explored solutions to overcome the dependence on the camera setup and parameters, it still remains a limitation to numerous existing approaches, particularly, for single-person solutions.

Therefore, future studies should focus on further improving the accuracy of 2D pose estimators as well as exploring methods that incorporate temporal consistency or depth information which have shown to be important factors in improving the performance for both single-person \cite{FusionFormer_2024, NakatsukaKomorita2022FewViews, YIN2024_infantpose} and multi-person approaches \cite{deFrançaSilva2022unsupervised_planesweep, WallisdeFrançaSilva2023607_UMVpose++, LinandLee_2021_PlaneSweep, Zhou2022ER-Net, Reddy2021tessetrack, Chu_2021_partaware, Zhu2023Crowded}. Also, multi-modal could help overcome some of the issues related to the poses mismatches or occlusions, since they can provide further information as input, such as depth information which would be much more precise than its estimation, or signal information like Wi-Fi or \gls{rf} which aren't affected by illumination conditions or occlusions.

Furthermore, the exploration of methods with lower levels of supervision has also, shown promising results for predicting the 3D pose with little to no annotated data \cite{deFrançaSilva2022unsupervised_planesweep, Liu2021dualview}. Therefore, future studies should explore this type of approach for either single or multi-person methodologies to try to develop a more effective model by, for instance, introducing a more effective loss function or synthesis of novel views, which would make it possible to be applied in different real-world scenarios whether labelled data is available or not. Also, a cost-effective method which almost hasn't been explored is the use of active learning, so the entire potential of this type of approach is yet to be assessed.

Finally, some researchers have also shown the potential benefits of selecting the best views instead of predicting the pose with all available views \cite{XiaZhang2022VitPose, Fan2021Multiagent}. By not using the information from all views, the quantity of information to process is also less, which would result in faster and more light-weighted models, which are two critical factors to have in consideration in order to have a model suitable to be deployed in the real-world.

%\subsection{Opportunities}

%\begin{figure}
%    \centering
%    \includegraphics[width=\linewidth]{images/summary1.png}
%    \caption{Summary of advantages and drawbacks of multi-view 3D pose estimation methodologies.}
%    \label{fig:summary}
%\end{figure}

%% file: sections/Challenges_and_Opportunities.tex
\section{Conclusion}\label{sec:conclusion}
In summary, most existing methods use a three-step methodology where first, the 2D pose is regressed from the input images, then, the 2D poses predicted for the different views are associated and finally, the 3D pose is constructed. However, this type of strategy can lead to unreliable 3D poses due to mismatches and erroneous 2D poses. Therefore, some methods have surpassed this problem by completely avoiding this step and working directly on the 3D space \cite{tu2020voxelpose}. Nevertheless, operations in the volumetric space become more computationally expensive, which might restrict their application in real-world settings. On the other hand, another methodology that has been utilised to tackle this problem is the selection of views. So, the model only uses, to predict the pose, the views that provide better visualisation of the human subjects. Consequently, by using just some of the available views, the computational burden is also reduced.

Another limitation that most proposed methods still have to face is the dependence on the camera setup and camera's calibration parameters, making most existing models unable to generalise to new viewpoints. To surpass this, researchers have exploited geometric multi-view consistency, stochasticity-based triangulation, adversarial learning and self-supervised learning methods.

Regarding the scarcity of 3D labelled datasets, methods that learn using a low level of supervision have been suggested to address this challenge. Nonetheless, some of those methods still depend on some 2D labelled data. On the other hand, the use of active learning has also, been suggested to overcome this issue. This strategy allows to dynamically acquire labels for unlabelled instances to boost model accuracy while requiring the least amount of annotation labour feasible. Nevertheless, only one work was found to employ this strategy, thus, more research is still required to determine the full potential of this technique for this task.

Even though many works have focused on solving this problem, the occurrence of occlusion still limits the model's performance in many cases. 
Temporal consistency has been widely used to track poses over time and also, to overcome the occlusion problem. Nonetheless, as pointed out by \cite{NakatsukaKomorita2022FewViews}, the use of this approach might suppress sudden and minor movements which can negatively impact the model's performance and also make it unsuitable for applications where the detection of those movements might be crucial, such as surveillance systems. Besides temporal consistency, the long-range dependencies between joints and the person, and the use of depth information have been proposed to improve the model's robustness against occlusions.

Finally, the use of more sophisticated types of cameras that can give additional information beyond RGB images has been shown to be able to provide reliable 3D pose estimations. These types of cameras have not yet been very explored, nonetheless, their employment could be advantageous because, according to the found works, the \gls{rgb-d}, \gls{tof} and 360º cameras, can directly provide depth information. By providing depth information directly as input, the computational cost associated with models using depth information can be substantially reduced, as the need for depth calculation will no longer be required. Regarding other types of signals, some researchers have explored the use of wireless signals to estimate the 3D pose. However, the association of information retrieved by wireless sensors with visual information given by cameras has not yet been investigated, leaving space for an innovative multi-modal approach combining both domains.

Thus, it's possible to conclude that, although various techniques have been proposed to try to overcome most challenges inherent to the estimation of the 3D pose, none of the existing approaches can balance high accuracy, quick inference, and low computational cost. So, the best model depends on the desired application, due to the trade-off between performance and complexity. Therefore, more research work is still needed in this area to find a suitable methodology for different application scenarios. The combination of a multi-modal approach with active learning may be the future for an efficient and accurate solution.

%% file: sections/Benchmarking.tex
%\newpage
\onecolumn
%\begin{strip}
\appendix
\section{Datasets - Benchmarking}\label{sec:benchmarking}

This section includes a summary of the results reported in the reviewed publications.

\subsection{Single-person 3D pose estimation}\label{app:singleperson}
Table \ref{tab:MPI-INF-3DHP}, Table \ref{tab:KTHMultiviewFootballII}, Table \ref{tab:HumanEva} and Table \ref{tab:totalcapture} present the results for the dataset MPI-INF-3DHP, KTH Multiview Football II, HumanEva and Total Capture, respectively.
%\end{strip}

\begin{table}[!htb]
\scriptsize
\centering
\caption{Performance of the revised methods in the MPI-INF-3DHP dataset}
%\begin{tabular}{m{1.45cm}M{0.5cm}M{1.2cm}M{1.2cm}M{1.65cm}}
\begin{tabular}{m{2cm}cccc}
\toprule
\textbf{Papers} & \textbf{Year} & \textbf{PCK (\%) $\uparrow$} & \textbf{AUC (\%) $\uparrow$} & \textbf{MPJPE (mm) $\downarrow$} \\ \midrule
\citet{kundu2020kinematicstructurepreserved} & \NoHyper\citeyear{kundu2020kinematicstructurepreserved} & \textbf{81.90} & \textbf{52.60} & 89.80 \\ %\midrule
\citet{Wang2022LHPEnetsAL} & \NoHyper\citeyear{Wang2022LHPEnetsAL} & --- & --- & 112.36 \\ %\midrule
\citet{WANG2022SmartVPoseNet} & \NoHyper\citeyear{WANG2022SmartVPoseNet} & --- & --- & 94.70 \\ %\midrule
\citet{Ma2023selfsupervised} & \NoHyper\citeyear{Ma2023selfsupervised} & 74.60 & 40.40 & --- \\ %\midrule
\citet{FusionFormer_2024} & \NoHyper\citeyear{FusionFormer_2024} & --- & --- & \textbf{5.40} \\ 
\bottomrule
\end{tabular}\label{tab:MPI-INF-3DHP}
\end{table}

%\onecolumn

\begin{table*}[!htb]
\scriptsize
\centering
\caption{Performance of the revised methods in the KTH Multiview Football II dataset}
\begin{tabular}{m{3.2cm}M{0.6cm}M{1.4cm}M{1.4cm}M{1.4cm}M{1.4cm}M{1.4cm}M{1.5cm}}
\toprule
\multirow{2.5}{*}{\textbf{Papers}} & \multirow{2.5}{*}{\textbf{Year}} & \multirow{2.5}{*}{\textbf{\# Cameras}} & \multicolumn{5}{c}{\textbf{PCP (\%) $\uparrow$}} \\ \cmidrule{4-8} 
 &  &  & \multicolumn{1}{c}{\textbf{Upper arms}} & \multicolumn{1}{c}{\textbf{Lower arms}} & \multicolumn{1}{c}{\textbf{Upper Legs}} & \multicolumn{1}{c}{\textbf{Lower Legs}} & \textbf{All parts (average)} \\ \midrule
\multirow{2}{*}{\citet{Belagiannis2014}} & \multirow{2}{*}{\NoHyper\citeyear{Belagiannis2014}} & 2 & \multicolumn{1}{c}{64.00} & \multicolumn{1}{c}{50.00} & \multicolumn{1}{c}{75.00} & \multicolumn{1}{c}{66.00} & 63.80 \\ 
 &  & 3 & \multicolumn{1}{c}{68.00} & \multicolumn{1}{c}{56.00} & \multicolumn{1}{c}{78.00} & \multicolumn{1}{c}{70.00} & 68.00 \\ \midrule
\multirow{2}{*}{\citet{belagiannis20163Dpami}} & \multirow{2}{*}{\NoHyper\citeyear{belagiannis20163Dpami}} & 2 & \multicolumn{1}{c}{96.00} & \multicolumn{1}{c}{68.00} & \multicolumn{1}{c}{98.00} & \multicolumn{1}{c}{88.00} & 87.50 \\ 
 &  & 3 & \multicolumn{1}{c}{98.00} & \multicolumn{1}{c}{72.00} & \multicolumn{1}{c}{99.00} & \multicolumn{1}{c}{92.00} & 90.30 \\ \midrule
\citet{ErshadiNasab2018MultipleH3} & \NoHyper\citeyear{ErshadiNasab2018MultipleH3} & 3 & \multicolumn{1}{c}{97.47} & \multicolumn{1}{c}{94.89} & \multicolumn{1}{c}{\textbf{100.00}} & \multicolumn{1}{c}{99.00} & 98.14 \\ \midrule
\citet{Ershadi-Nasab2021Adversariallearning} & \NoHyper\citeyear{Ershadi-Nasab2021Adversariallearning} & 3 & \multicolumn{1}{c}{\textbf{100.00}} & \multicolumn{1}{c}{\textbf{99.60}} & \multicolumn{1}{c}{\textbf{100.00}} & \multicolumn{1}{c}{\textbf{99.60}} & \textbf{99.80} \\ \bottomrule
\end{tabular}\label{tab:KTHMultiviewFootballII}
\end{table*}

\begin{table*}[!htb]
\scriptsize
\centering
\caption{3D error (mm) $\downarrow$ of the revised methods in the HumanEva dataset}
\begin{tabular}{lccccccccccc}
\toprule
\multirow{2}{*}{\textbf{Papers}} & \multirow{2}{*}{\textbf{Year}} & \multicolumn{4}{c}{\textbf{Sequence   1}} & \multicolumn{3}{c}{\textbf{Sequence 2}} & \multicolumn{3}{c}{\textbf{Sequence 3}}\\ \cmidrule(r){3-6} \cmidrule(r){7-9} \cmidrule(r){10-12}
 &  & \multicolumn{1}{c}{\textbf{Walk}} & \multicolumn{1}{c}{\textbf{Jog}} & \multicolumn{1}{c}{\textbf{Balance}} & \textbf{Box} & \multicolumn{1}{c}{\textbf{Walk}} & \multicolumn{1}{c}{\textbf{Jog}} & \textbf{Balance} & \multicolumn{1}{c}{\textbf{Walk}} & \multicolumn{1}{c}{\textbf{Jog}} & \textbf{Combo}\\ \midrule
\citet{humaneva} & \NoHyper\citeyear{humaneva}\endNoHyper & \multicolumn{1}{c}{76.0} & \multicolumn{1}{c}{85.0} & \multicolumn{1}{c}{86.0} & --- & \multicolumn{1}{c}{60.0} & \multicolumn{1}{c}{93.0} & 80.0 & \multicolumn{1}{c}{---} & \multicolumn{1}{c}{---} & --- \\ %\midrule
\citet{Amin2013MultiviewPS} & \NoHyper\citeyear{Amin2013MultiviewPS}\endNoHyper & \multicolumn{1}{c}{54.5} & \multicolumn{1}{c}{---} & \multicolumn{1}{c}{---} & \textbf{47.7} & \multicolumn{1}{c}{50.2} & \multicolumn{1}{c}{---} & --- & \multicolumn{1}{c}{54.7} & \multicolumn{1}{c}{54.0} & \textbf{51.8} \\ %\midrule
\citet{Belagiannis2014} & \NoHyper\citeyear{Belagiannis2014}\endNoHyper & \multicolumn{1}{c}{68.3} & \multicolumn{1}{c}{---} & \multicolumn{1}{c}{---} & 62.7 & \multicolumn{1}{c}{---} & \multicolumn{1}{c}{---} & --- & \multicolumn{1}{c}{---} & \multicolumn{1}{c}{---} & --- \\ %\midrule
\citet{belagiannis20163Dpami} & \NoHyper\citeyear{belagiannis20163Dpami}\endNoHyper & \multicolumn{1}{c}{68.3} & \multicolumn{1}{c}{---} & \multicolumn{1}{c}{---} & 62.7 & \multicolumn{1}{c}{---} & \multicolumn{1}{c}{---} & --- & \multicolumn{1}{c}{---} & \multicolumn{1}{c}{---} & --- \\ %\midrule
\citet{mehrizi2018markerfree} & \NoHyper\citeyear{mehrizi2018markerfree}\endNoHyper & \multicolumn{1}{c}{\textbf{40.4}} & \multicolumn{1}{c}{\textbf{43.0}} & \multicolumn{1}{c}{---} & --- & \multicolumn{1}{c}{\textbf{23.5}} & \multicolumn{1}{c}{\textbf{45.1}} & --- & \multicolumn{1}{c}{\textbf{33.4}} & \multicolumn{1}{c}{\textbf{30.9}} & --- \\
\bottomrule
\end{tabular}\label{tab:HumanEva}
\end{table*}

\begin{table*}[!htb]
\scriptsize
\centering
\caption{MPJPE (mm) $\downarrow$ of the revised methods in the Total Capture dataset (Avg. - Average).}
\begin{tabular}{m{1.75cm}M{0.5cm}M{0.6cm}M{0.5cm}M{0.5cm}M{0.5cm}M{0.5cm}M{0.5cm}M{0.6cm}M{0.5cm}M{0.5cm}M{0.5cm}M{0.5cm}M{0.5cm}M{0.5cm}M{0.6cm}}
\toprule
\multirow{3}{*}{\textbf{Papers}} & \multirow{3}{*}{\textbf{Year}} & \multicolumn{7}{c}{\textbf{Seen Cameras (1, 3, 5, 7)}} & \multicolumn{7}{c}{\textbf{Unseen Cameras (2, 4, 6, 8)}} \\ \cmidrule(r){3-9} \cmidrule(r){10-16}
 &   & \multicolumn{3}{c}{\textbf{Seen Subject 1, 2, 3}} & \multicolumn{3}{c}{\textbf{Unseen Subject 4, 5}} & \multirow{2}{*}{\textbf{Avg.}} & \multicolumn{3}{c}{\textbf{Seen Subject 1, 2, 3}} & \multicolumn{3}{c}{\textbf{Unseen Subject 4, 5}} & \multirow{2}{*}{\textbf{Avg.}} \\ \cmidrule(r){3-5} \cmidrule(r){6-8} \cmidrule(r){10-12} \cmidrule(r){13-15}
  & \multicolumn{1}{c}{} & \textbf{W2} & \textbf{FS3} & \textbf{A3} & \textbf{W2} & \textbf{FS3} & \textbf{A3} &  & \textbf{W2} & \textbf{FS3} & \textbf{A3} & \textbf{W2} & \textbf{FS3} & \textbf{A3} &  \\ \midrule
\citet{remelli2020lightweight} & \NoHyper\citeyear{remelli2020lightweight}\endNoHyper & 10.60 & 30.40 & 16.30 & 27.00 & 65.00 & 34.20 & 27.50 & 22.40 & 47.10 & 27.80 & 39.10 & 75.70 & 43.10 & 38.20 \\%\midrule
\citet{WAN2023_Viewconsistency} & \NoHyper\citeyear{WAN2023_Viewconsistency}\endNoHyper & 13.00 & 24.00 & 17.00 & 23.00 & 41.00 & 29.00 & 23.00 & --- & --- & --- & --- & --- & --- & --- \\%\midrule
\citet{FusionFormer_2024} & \NoHyper\citeyear{FusionFormer_2024}\endNoHyper & \textbf{5.50} & \textbf{15.00} & \textbf{5.68} & \textbf{18.10} & \textbf{37.60} & \textbf{20.60} & \textbf{15.00} & \textbf{22.10} & \textbf{35.40} & \textbf{23.40} & \textbf{23.20} & \textbf{42.60} & \textbf{28.40} & \textbf{28.30} \\ \midrule \midrule
\multicolumn{16}{p{14cm}}{Avg. - Average \newline \textbf{Testing sequences:} W2 - Walking2, FS3 - Freestyle3, A3 - Acting3.
\newline\textbf{Training sequences:} ROM1,2,3; Walking1,3; Freestyle1,2; Acting1,2 and Running1 using subjects 1, 2 and 3.}\\\bottomrule
\end{tabular}\label{tab:totalcapture}
\end{table*}
%\break\hfill

%\onecolumn
\newpage
\subsection{Multi-person 3D pose estimation}\label{app:multiperson}
Table \ref{tab:UMPM} and Table \ref{tab:CMUPanoptic} present the results for the dataset \gls{umpm} and CMU Panoptic, respectively.

\begin{table*}[!htb]
\scriptsize
\centering
\caption{Performance of the revised methods in the \gls{umpm} dataset}
\begin{tabular}{m{2.5cm}M{0.8cm}M{0.9cm}M{1.1cm}M{1.1cm}M{1.1cm}M{1.1cm}M{1.1cm}M{1.1cm}M{1.1cm}}
\toprule
\multirow{3.5}{*}{\textbf{Papers}} & \multicolumn{2}{c}{\multirow{3.5}{*}{\textbf{Year}}} & \multicolumn{7}{c}{\textbf{PCP (\%) $\uparrow$}} \\ \cmidrule{4-10} 
 & \multicolumn{2}{c}{} & \textbf{Head} & \textbf{Torso} & \textbf{Upper arms} & \textbf{Lower arms} & \textbf{Upper Legs} & \textbf{Lower Legs} & \textbf{Average} \\ \midrule
\citet{ErshadiNasab2018MultipleH3} & \multicolumn{2}{c}{\NoHyper\citeyear{ErshadiNasab2018MultipleH3}} & 95.87 & 98.53 & 92.42 & 84.92 & 94.23 & 86.36 & 91.02 \\ %\midrule
\citet{Ershadi-Nasab2021Adversariallearning} & \multicolumn{2}{c}{\NoHyper\citeyear{Ershadi-Nasab2021Adversariallearning}} & \textbf{99.40} & \textbf{99.70} & \textbf{95.30} & \textbf{89.80} & \textbf{96.60} & \textbf{90.40} & \textbf{94.30} \\ %\midrule
\midrule
\multirow{3.5}{*}{\textbf{Papers}} & \multirow{3.5}{*}{\textbf{Year}} & \multirow{3.5}{*}{\textbf{Actors}} & \multicolumn{7}{c}{\textbf{UMPM sequence: p3-chair-11 --- PCP (\%) $\uparrow$}} \\ \cmidrule{4-10} 
 & \multicolumn{2}{c}{} & \textbf{Head} & \textbf{Torso} & \textbf{Upper arms} & \textbf{Lower arms} & \textbf{Upper Legs} & \textbf{Lower Legs} & \textbf{Average} \\ \midrule
\multirow{1}{*}{{\NoHyper\citeauthor{ErshadiNasab2018MultipleH3}}} & \multirow{2}{*}{\NoHyper\citeyear{ErshadiNasab2018MultipleH3}} & A1 & 94.76 & 96.82 & 91.24 & 83.78 & 92.45 & 85.56 & 89.76 \\  
\cite{ErshadiNasab2018MultipleH3} &  & A2 & 95.87 & 97.28 & 95.12 & 83.89 & 93.76 & 85.12 & 89.89 \\ \midrule
\multirow{1}{*}{{\NoHyper\citeauthor{Ershadi-Nasab2021Adversariallearning}}} & \multirow{2}{*}{\NoHyper\citeyear{Ershadi-Nasab2021Adversariallearning}} & A1 & \textbf{99.42} & \textbf{99.34} & \textbf{98.89} & \textbf{91.56} & \textbf{97.11} & \textbf{96.32} & \textbf{96.65} \\  
\cite{Ershadi-Nasab2021Adversariallearning} &  & A2 & \textbf{98.31} & \textbf{99.20} & \textbf{99.10} & \textbf{90.22} & \textbf{96.45} & \textbf{94.65} & \textbf{95.83} \\ \bottomrule
\end{tabular}\label{tab:UMPM}
\end{table*}

\begin{table*}[!htb]
\scriptsize
\centering
\caption{Performance of the revised methods in the CMU Panoptic dataset}
\begin{tabular}{M{0.8cm}m{2.95cm}M{0.6cm}M{1cm}M{0.9cm}M{0.9cm}M{0.9cm}M{0.9cm}M{0.9cm}M{0.9cm}M{1cm}} 
\toprule
\textbf{\#Views} & \textbf{Paper} & \textbf{Year} & \textbf{MPJPE (mm) $\downarrow$} & \textbf{AP$_{25}\uparrow$} & \textbf{AP$_{50}\uparrow$} & \textbf{AP$_{75}\uparrow$} & \textbf{AP$_{100}\uparrow$} & \textbf{AP$_{125}\uparrow$} & \textbf{AP$_{150}\uparrow$} & \textbf{mAP $\uparrow$} \\ \midrule
\multirow{3}{*}{2} & \citet{wang2021direct} & 2021 & 34.80 & 37.70 & --- & --- & 93.00 & --- & --- & --- \\ %\cmidrule{2-11} 
 & \citet{Zhu2023Crowded} & 2023 & 47.42 & 25.51 & 62.03 & 77.89 & 86.14 & 90.16 & 92.39 & 72.35 \\ %\cmidrule{2-11} 
 & \citet{Zhuang20231763_FasterVoxelPose+} & 2023 & 42.53 & 31.25 & 65.63 & --- & 93.51 & --- & 96.12 & --- \\ \midrule
\multirow{6}{*}{3} & \citet{tu2020voxelpose} & 2020 & 24.29 & 58.94 & 93.88 & --- & 98.45 & --- & 99.32 & --- \\ %\cmidrule{2-11} 
 & \citet{wang2021direct} & 2021 & 21.10 & 71.80 & --- & --- & 95.10 & --- & --- & --- \\ %\cmidrule{2-11} 
 & \citet{zhang2021voxeltrack} & 2021 & 24.93 & 49.09 & 92.44 & --- & 97.62 & --- & --- & --- \\ %\cmidrule{2-11} 
 & \citet{ye2022fastervoxelpose} & 2022 & 26.13 & 53.68 & 91.89 & --- & 97.40 & --- & 98.30 & --- \\ %\cmidrule{2-11}
 & \citet{Zhu2023Crowded} & 2023 & 33.03 & 34.98 & 76.72 & 89.37 & 93.87 & 96.06 & 97.24 & 81.37 \\ %\cmidrule{2-11} 
 & \citet{Zhuang20231763_FasterVoxelPose+} & 2023 & 24.98 & 57.23 & 92.21 & --- & 97.83 & --- & 98.32 & --- \\ \midrule
\multirow{6}{*}{4} & \citet{Fan2021Multiagent} & 2021 & 113.60 & --- & --- & --- & --- & --- & --- & --- \\ %\cmidrule{2-11} 
 & \citet{wang2021direct} & 2021 & 19.30 & 84.10 & --- & --- & 96.70 & --- & --- & --- \\ %\cmidrule{2-11} 
 & \citet{zhang2021voxeltrack} & 2021 & 20.35 & 66.20 & 96.34 & --- & 99.47 & --- & --- & --- \\ %\cmidrule{2-11} 
 & \citet{ye2022fastervoxelpose} & 2022 & 21.12 & 73.95 & 97.02 & --- & 99.21 & --- & 99.35 & --- \\ %\cmidrule{2-11} 
 & \citet{Zhu2023Crowded} & 2023 & 23.89 & 52.84 & 90.80 & 96.63 & 97.85 & 98.46 & 98.88 & 89.24 \\ %\cmidrule{2-11}
  & \citet{Zhuang20231763_FasterVoxelPose+} & 2023 & 20.95 & 75.92 & 97.86 & --- & 99.32 & --- & 99.40 & --- \\ \midrule
\multirow{10}{*}{5} & \citet{tu2020voxelpose} & 2020 & 17.68 & 83.59 & 98.33 & --- & 99.76 & --- & \textbf{99.91} & --- \\ %\cmidrule{2-11} 
 & \citet{Fan2021Multiagent} & 2021 & 94.21 & --- & --- & --- & --- & --- & --- & --- \\ %\cmidrule{2-11} 
 & \citet{LinandLee_2021_PlaneSweep} & 2021 & 16.75 & 92.12 & \textbf{98.96} & \textbf{} & \textbf{99.81} & \textbf{} & 99.84 & --- \\ %\cmidrule{2-11} 
 & \citet{Reddy2021tessetrack} & 2021 & \textbf{7.30} & --- & --- & --- & --- & --- & --- & --- \\ %\cmidrule{2-11} 
 & \citet{wang2021direct} & 2021 & 15.80 & \textbf{92.30} & 96.60 & --- & 97.50 & --- & 97.70 & --- \\ %\cmidrule{2-11}  
 & \citet{zhang2021voxeltrack} & 2021 & 16.97 & 85.88 & 98.31 & --- & 99.54 & --- & --- & --- \\ %\cmidrule{2-11} 
 & \citet{ye2022fastervoxelpose} & 2022 & 18.26 & 85.22 & 98.08 & --- & 99.32 & --- & 99.48 & --- \\ %\cmidrule{2-11}
 & \citet{Chen2023VTP} & 2023 & 17.62 & 83.79 & 97.14 & --- & 98.15 & --- & 98.40 & --- \\ %\cmidrule{2-11} 
 & \citet{Zhu2023Crowded} & 2023 & 18.88 & 61.28 & 95.10 & \textbf{98.70} & 99.39 & \textbf{99.76} & 99.87 & \textbf{92.35} \\ %\cmidrule{2-11} 
 & \citet{Zhuang20231763_FasterVoxelPose+} & 2023 & 17.42 & 86.25 & 98.45 & --- & 99.77 & --- & 99.82 & --- \\ \midrule
6 & \citet{Fan2021Multiagent} & 2021 & 87.53 & --- & --- & --- & --- & --- & --- & --- \\ \midrule
7 & \citet{Fan2021Multiagent} & 2021 & 84.02 & --- & --- & --- & --- & --- & --- & --- \\ \midrule
8 & \citet{Fan2021Multiagent} & 2021 & 80.81 & --- & --- & --- & --- & --- & --- & --- \\ \midrule \midrule
\multicolumn{11}{p{16cm}}{
Sequences used by \citet{tu2020voxelpose, Reddy2021tessetrack, ye2022fastervoxelpose, zhang2021voxeltrack, Chen2023VTP, LinandLee_2021_PlaneSweep, Zhuang20231763_FasterVoxelPose+} \newline
\textbf{Train set:} "160422\_ultimatum1", "160224\_haggling1", "160226\_haggling1", "161202\_haggling1", "160906\_ian1", "160906\_ian2", "160906\_ian3", "160906\_band1", "160906\_band2" and "160906\_band3".
\newline \textbf{Test set:} "160906\_pizza1", "160422\_haggling1", "160906\_ian5" and "160906\_band4". \hfill \break
\newline \citet{wang2021direct} -- training set includes the same sequences presented above except for "160906\_band3". The test sequences are the same.
\newline \citet{Fan2021Multiagent} -- training set: "Mafia", validation set: "Mafia" and "Ultimatum" and testing set: "Mafia", "Ultimatum" and "Haggling".
\newline \citet{Zhu2023Crowded} -- training set: "160422\_ultimatum1" and testing set: "160906\_pizza1".
}\\ \bottomrule
\end{tabular}\label{tab:CMUPanoptic}
\end{table*}

\twocolumn